\documentclass{article}

\usepackage{microtype}
\usepackage{graphicx}
\usepackage{subcaption}
\usepackage{booktabs} 
\usepackage{multirow}
\usepackage{amsfonts}
\usepackage{nicefrac}
\usepackage{xcolor}
\usepackage{url}

\usepackage{hyperref}

\usepackage[preprint]{icml2026}

\usepackage{amsmath}
\usepackage{amssymb}
\usepackage{mathtools}
\usepackage{amsthm}

\usepackage[capitalize,noabbrev]{cleveref}

\theoremstyle{plain}

\theoremstyle{definition}

\theoremstyle{remark}

\icmltitlerunning{IKNO: Infinite-order Kernel Neural Operators}

\begin{document}

\twocolumn[
  \icmltitle{IKNO: Infinite-order Kernel Neural Operators}

  \begin{icmlauthorlist}
    \icmlauthor{Pengyuan Zhu}{ntu}
    \icmlauthor{Ivor W. Tsang}{astar,ntu}
    \icmlauthor{Yueming Lyu}{astar}
  \end{icmlauthorlist}

  \icmlaffiliation{ntu}{Nanyang Technological University}
  \icmlaffiliation{astar}{Centre for Frontier AI Research(CFAR), Agency for Science, Technology and Research (A*STAR)}

 \icmlcorrespondingauthor{Yueming Lyu}{Lyu\_Yueming@a-star.edu.sg}

  \icmlkeywords{Neural Operators, Kernel Methods, Scientific Machine Learning, PDEs}

  \vskip 0.3in
]

\printAffiliationsAndNotice{}

\begin{abstract}
Neural operators have achieved significant success in modern scientific computing due to their flexibility and strong generalization capabilities. Existing models, however, primarily rely on first-order kernel integral approximations, which severely limit their expressivity. To address this, we propose the Infinite-order Kernel Neural Operator (IKNO), which constructs neural operators via infinite-order kernel integrals and admits an elegant closed-form finite approximation.
We develop two complementary infinite-order neural operator constructions: IKNO-Vanilla, which applies the full-kernel resolvent on the product grid via Kronecker eigendecomposition, and IKNO-TP, an alternative tensor-product operator that composes per-axis resolvents. Furthermore, we develop fast computation schemes for both variants of IKNO, which achieve outstanding global information aggregation while maintaining high computational efficiency. 
Empirically, we evaluate our IKNO on both time-dependent and time-independent benchmarks with arbitrary input shapes, including large-scale industrial datasets. Extensive experiments demonstrate that the IKNO method consistently achieves the SOTA accuracy with significant improvements on nearly all benchmark datasets while maintaining scalability to very large point clouds. 
\end{abstract}


\section{Introduction}

Partial differential equations serve as fundamental tools for describing how properties evolve over time and space, finding applications across physics, finance, and numerous other fields\cite{evans2022partial}. For instance, weather and sea condition forecasts rely on solving the NS equations\cite{vallis2017atmospheric}. In finance, pricing many financial products requires solving the Black-Scholes equation\cite{black1973pricing}. In medicine, partial differential equations are used to describe drug diffusion within tissues\cite{murray2007mathematical}. 

Traditional methods for solving partial differential equations involve numerical simulation techniques such as FEM\cite{hughes2003finite}, FDM\cite{thomas2013numerical}, and others. The issue with these methods lies in their relatively slow computational speed. Particularly in industrial design, computational complexity increases exponentially\cite{hughes2003finite} as mesh resolution is refined, making high-fidelity simulations prohibitively expensive. Furthermore, these approaches can only determine performance metrics after a shape is designed. However, the goal of industrial design is to derive shapes with better performance. Traditional PDE solvers fall far short of meeting the demands of the industrial design field.

In recent years, to overcome the limitations of traditional numerical methods in computational efficiency and generalization capability, data-driven deep learning PDE solvers have emerged. Within this field, Neural Operators\cite{li2020fourier} stand out as a highly promising paradigm. Unlike conventional deep learning models, neural operators aim to learn nonlinear mappings between infinite-dimensional function spaces, thereby offering a novel approach to achieving efficient and high-precision PDE solutions.

Although existing neural operator models achieve considerable progress in solving PDEs on arbitrary point clouds,  they primarily rely on first-order kernel integral approximations\cite{li2020neural}\cite{li2023geometry}\cite{wen2025geometry}\cite{brandstetter2022message}, which severely limit their expressivity. Our preliminary experiment on \textit{Poisson-C-Sines} shows a clear monotonic decrease in error as the finite-order kernel propagation order increases from 1 to 4 (Table~\ref{tab:finite_order_motivation}). This preliminary experiment demonstrates that higher-order kernel integral propagation can aggregate and summarize global information more effectively, leading to better representations and enhanced model performance. These insights suggest that neural operators may be significantly enhanced by incorporating higher-order kernel integrals. Consequently, we investigate a pivotal question: \textit{can we formulate an infinite-order kernel operator to achieve comprehensive information aggregation over all propagation orders?}    To answer this question, we propose a novel neural operator model, the Infinite-order Kernel Neural Operator (IKNO), which improves expressivity via infinite-order kernel integrals that benefit from superior global information aggregation and propagation. 

Our main contributions are summarized as follows:
\begin{itemize}
 \item  We propose a novel Infinite-order Kernel Neural Operator
(IKNO), which constructs neural operators via
infinite-order kernel integrals that  benefit from superior global information aggregation and propagation. Moreover, we derive an elegant closed-form finite approximation for the infinite-order kernel integral operators. 

\item We introduce two complementary multi-dimensional infinite-order neural operator constructions: IKNO-Vanilla, which applies the full-kernel resolvent on the product grid via Kronecker eigendecomposition, and IKNO-TP, an alternative tensor-product operator that composes per-axis resolvents and thereby induces an implicit functional regularization. These two constructions are not algebraically identical, but both admit Kronecker-structured fast computation that reduces the dominant preprocessing cost from $O(N^{3d})$ to $O(d \cdot N^{3})$ with essentially identical wall-clock cost, enabling near-instantaneous computation of global dependencies on high-resolution grids.

\item Empirically, our IKNO method consistently achieves state-of-the-art accuracy with significant improvements over the previous SOTA (GAOT~\cite{wen2025geometry}) on the large majority of PDE benchmarks while maintaining scalability to very large point clouds. Across 15 benchmarks, the two variants exhibit complementary strengths, with IKNO-TP providing the strongest overall performance and IKNO-Vanilla offering a competitive non-tensorized alternative across diverse PDE settings.

\end{itemize}


\section{Related work}

Early studies on neural operator learning primarily focused on regular grids. The most representative among them is the Fourier Neural Operator\cite{li2020fourier}, which achieves exceptional computational efficiency by parameterizing integral kernels in the frequency domain and utilizing FFT. Furthermore, to better capture features across different spatial scales in physical fields, U-NO\cite{rahman2022u} with multi-scale collaborative structure of U-Net was proposed. However, due to their underlying design, these models are highly dependent on regular grids. Consequently, they are only applicable to uniformly discretized regular grid datasets, which greatly limits their application in complex geometric domains.

Subsequently, researchers sought to extend the advantages of neural operator to non-regular spaces and larger-scale datasets to achieve better generalization and practicality. However, when addressing practical engineering problems, the input data mesh often comprises a vast number of points. Learning Neural Operator directly in the original high-resolution space would bring memory overhead and computational complexity. To address this challenge, recent neural operator models typically adopt a strategy of mapping large-scale unstructured data into a compact latent space via an encoder\cite{wang2024latent,wu2024transolver,luo2025transolver++}. By decoupling inputs' resolution from model's internal resolution, the trained neural operators can maintain strong generalization capabilities across varying resolutions and sampling methods. Based on the different methods of latent space representation, two mainstream approaches have emerged in research area.

\subsection{Latent grid-based mapping}
The core idea of this approach is to encode unstructured input data onto a regular latent grid. The regular grid inherently preserves geometric information, enabling the model to implicitly encode spatial structure. Operators dependent on the grid can also be directly applied in the latent space, such as convolutions, Fourier transforms, and multi-scale operators. Geo-FNO\cite{li2023fourier} maps input points into a regular grid via a reversible transformation, successfully extending the FNO\cite{li2020fourier} model to irregular domains. GINO\cite{li2023geometry} achieves mapping from physical coordinates to a latent grid space using a learnable first-order kernel function, then processes this grid space using the FNO\cite{li2020fourier} model architecture. Li et al.~\cite{li2025geometric} employ optimal transport to map inputs and outputs onto a latent mesh space, enabling efficient neural operator learning within this latent domain. GAOT\cite{wen2025geometry} utilizes a multi-scale encoder from GINO\cite{li2023geometry} with learnable weights, supplemented by artificially designed geometric embeddings, to map inputs onto the latent grid space. Within this space, a ViT\cite{dosovitskiy2020image} model is employed to efficiently learn the latent neural operator.

\subsection{Latent token-based mapping}
Latent token-based methods no longer explicitly map input functions to a regular grid. Instead, they employ learnable encoding mechanisms to compress raw inputs into a finite set of implicit token representations. These tokens serve as low-dimensional summaries of the input physics information or key structural features, regarded as fundamental units for subsequent neural operator computations. The encoder component has few constraints on the mapping, resulting in strong expressive capabilities, and it can be compatible with various geometric input formats such as point clouds, SDF\cite{park2019deepsdf}, parametric curves, and meshes.

LNO\cite{wang2024latent} maps input coordinates and physical quantities to latent tokens via an encoder centered on a cross-attention\cite{vaswani2017attention} module. Transolver\cite{wu2024transolver} weights all information through an MLP\cite{rumelhart1986learning} to map onto latent slices, incorporating a PhysicsAttention mechanism designed to aggregate information from multiple slices. Transolver++\cite{luo2025transolver++} further enhances this by adding constraints to the weighted mapping, enabling each slice to carry more diverse information and boosting the encoder's expressive ability. Both OFermer\cite{li2022transformer} and GNOT\cite{hao2023gnot} directly map input physical quantities to tokens via MLP. The former employs a special attention module resembling kernel functions, offering both Fourier-type and Galerkin-type\cite{galerkin1968rods} variants. The latter utilizes a specially designed Heterogeneous Normalized Cross-Attention module to encode and incorporate diverse input forms, such as shape and edge information, into the latent neural operator.

However, current latent grid-based methods lack sufficient expressive power, often requiring manual design of kernel function summation ranges\cite{wen2025geometry}. This leads to the loss of information at different scales and over long distances. Meanwhile, latent token-based methods struggle to learn geometric structural information. Too few constraints also make them overly sensitive to hyperparameters, prone to overfitting\cite{luo2025transolver++}. To deal with these problems, inspired by the theory of integral equations, we introduce the infinite-order kernel integral. Instead of a single aggregation step, we consider the multiplier of all higher-order interactions. This approach allows the model to aggregate information across the entire input domain more effectively while simultaneously capturing geometric structure features.

\begin{figure*}[!t]
\centering
\includegraphics[width=0.9\linewidth]{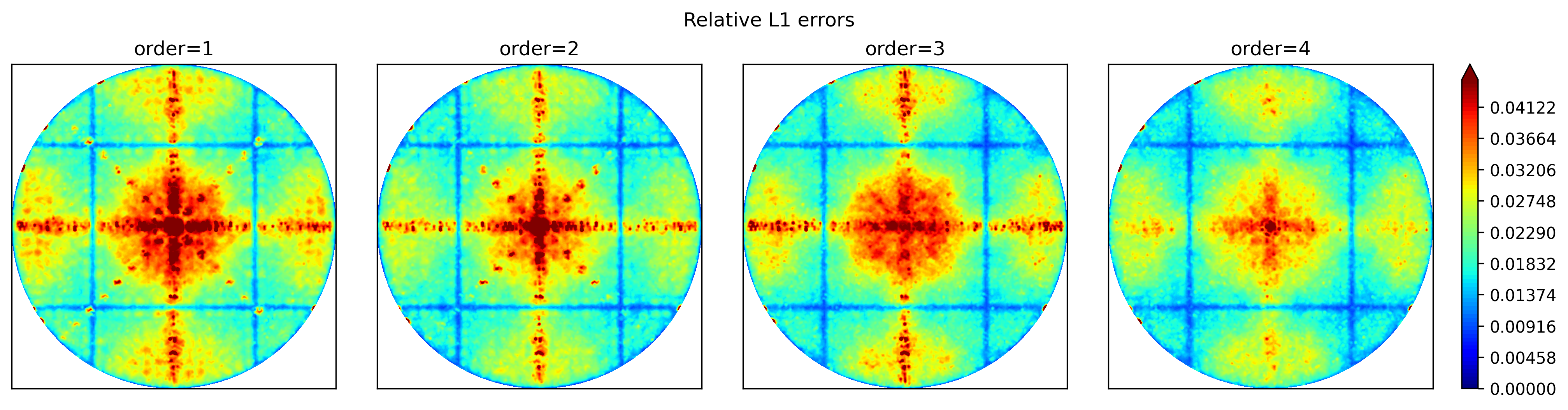}
\caption{Visualization of relative $L_1$ error fields for finite-order kernel propagation on the \textit{Poisson-C-Sines} benchmark. As the propagation order increases from $p=1$ to $p=4$, both the error magnitude and the spatial extent of high-error regions decrease.}
\label{fig:finite_order_error_fields}
\end{figure*}


\section{Methodology}
\paragraph{Problem setup.}
We study neural operator learning for PDE solution maps on arbitrary point clouds. For time-independent problems, the model learns an operator $\mathcal{G}: \mathcal{A}\to\mathcal{U}$ mapping input conditions $a$ to solutions $u$ on $\Omega$. For time-dependent problems, it learns a trajectory operator from the initial state and time-varying conditions to the solution trajectory over $\Omega_T=[0,T]$. In all cases, inputs and outputs are observed on point clouds; the detailed formulation is provided in Appendix~\ref{app:problem_formulation}.

\subsection{Motivation: from finite-order to infinite-order kernel propagation}
\label{sec:motivation_finite_order}

Many recent neural operators for irregular or large-scale domains follow an encoder--processor--decoder structure~\cite{li2023geometry,wu2024transolver,wen2025geometry,mousavi2025rigno}: the encoder maps observations from the physical input domain into a latent representation, the processor performs global mixing in the latent space, and the decoder maps the processed latent representation back to arbitrary query locations. In a preliminary finite-order study on the \textit{Poisson-C-Sines} benchmark, we instantiate this paradigm with a fixed latent grid $G=G_1\times\cdots\times G_d$ that has the same spatial dimensionality as the input domain. Each axis contains $L$ grid points, so the latent grid contains $L^d$ points independent of the input point-cloud resolution.

We replace the conventional first-order kernel aggregation with a truncated kernel propagation operator on the latent grid,
\begin{equation}
    \boldsymbol{R}_{p}(G) =
    \boldsymbol{I} + \alpha \boldsymbol{K}_{GG} + \cdots + \alpha^p \boldsymbol{K}_{GG}^{p},
\end{equation}
where $\boldsymbol{K}_{GG}\in\mathbb{R}^{L^d\times L^d}$ is the kernel Gram matrix over latent grid points and $p$ is the propagation order. In this preliminary study, we use a fixed linear-window kernel $k(\boldsymbol{x},\boldsymbol{y})=\max(1-\|\boldsymbol{x}-\boldsymbol{y}\|/r,0)$ with radius $r=0.2$ and unit scale. Given input points $P=\{\boldsymbol{x}^i\}_{i=1}^{\bar{N}}$ with token matrix $\boldsymbol{V}_{P}$ and query points $Q=\{\boldsymbol{q}^r\}_{r=1}^{N_q}$, the finite-order encoder--processor--decoder pipeline is
\begin{align}
    \boldsymbol{V}_{G} &= \boldsymbol{R}_{p}(G)\boldsymbol{K}_{GP}\boldsymbol{V}_{P}, \\
    \boldsymbol{V}'_{G} &= \mathrm{TransformerEncoder}(\boldsymbol{V}_{G}), \\
    \boldsymbol{V}_{Q} &= \boldsymbol{K}_{QG}\boldsymbol{R}_{p}(G)\boldsymbol{V}'_{G},
\end{align}
where $\boldsymbol{K}_{GP}$ maps input points to the latent grid and $\boldsymbol{K}_{QG}$ maps latent-grid features to query points. We fix the propagation coefficient to $\alpha=-0.15$ and keep both the kernel scale and $\alpha$ non-learnable; additional notation details are provided in Appendix~\ref{app:finite_order_motivation}.

The quantitative results are summarized in Table~\ref{tab:finite_order_motivation}, and the corresponding error-field visualizations are shown in Fig.~\ref{fig:finite_order_error_fields}. The aggregate error decreases monotonically as the propagation order increases from $p=1$ to $p=4$, dropping from 2.49 to 2.13. The visualizations further show that higher orders not only reduce the error magnitude, but also shrink the spatial regions with large relative error. This controlled trend provides direct empirical evidence that higher-order kernel propagation improves global information aggregation. Rather than treating the truncation order as a manually selected hyperparameter, the infinite-order formulation below aggregates all propagation orders through a closed-form operator.




\begingroup
\setlength{\intextsep}{2pt}
\begin{table}[!h]
\centering
\captionsetup{font=scriptsize,skip=1pt}
\caption{Preliminary finite-order kernel results on the \textit{Poisson-C-Sines} benchmark using median relative $L_1$ error ($\%$), with a fixed linear-window kernel ($r=0.2$, unit scale, $\alpha=-0.15$).}
\label{tab:finite_order_motivation}
\begin{tabular}{lc}
\toprule
\textbf{Order $p$} & \textbf{Median Relative $L_1$ Error ($\%$)} \\
\midrule
1 & 2.49 \\
2 & 2.32 \\
3 & 2.25 \\
4 & 2.13 \\
\bottomrule
\end{tabular}
\end{table}
\endgroup

\subsection{Infinite-order kernel integral}

\subsubsection{Definitions and discretization}
One fundamental component of a Neural Operator is the first-order kernel integral operator, which computes embedding aggregation by considering global information:
\begin{equation}
\label{Kintegral}
\mathcal{I}[v](\boldsymbol{x})=\int_{\mathcal{X}}k(\boldsymbol{x},\boldsymbol{y})v(\boldsymbol{y}) \text{d} \boldsymbol{y}
\end{equation}
where $k(\boldsymbol{x},\boldsymbol{y})$ denotes the kernel function with input $\boldsymbol{x},\boldsymbol{y} \in \mathcal{X} \subseteq \mathbb{R}^d$, and $v(\cdot) : \mathcal{X} \rightarrow \mathbb{R}^h  $  denotes feature map function.

In practical applications, the continuous integral in Eq.(\ref{Kintegral}) can be computed via a discrete approximation over $N$ samples:
\begin{equation}
\hat{\mathcal{I}}_{1}[v](\boldsymbol{x})=\frac{1}{N}\sum_{i=1}^{N}k(\boldsymbol{x},\boldsymbol{y}^{i})v(\boldsymbol{y}^{i})
\end{equation}
where $\boldsymbol{y}^i$ denotes the $i$-th sample of variable $\boldsymbol{y}$. 

To achieve superior global information aggregation and propagation, we introduce   the $n$-order kernel integral (with $n \ge 2$)  as follows:
\begin{equation}
\begin{aligned}
\mathcal{I}_{n}[v](\boldsymbol{x})
\;=\;& \int_{\mathcal{X}} \dots \int_{\mathcal{X}}
k(\boldsymbol{x},\boldsymbol{y}_{1}) \times \dots \\
&\times k(\boldsymbol{y}_{n-1},\boldsymbol{y}_n)
\times v(\boldsymbol{y}_n)
\text{d}\boldsymbol{y}_1  \dots \text{d}\boldsymbol{y}_n
\end{aligned}
\end{equation}
where $\boldsymbol{y}_m \in \mathbb{R}^d$ denotes the $m$-th vector variable for integration. The kernel $k(\cdot,\cdot)$ is chosen to ensure the multi-integral exists. For example, given a bounded $v(\cdot)$ and $\forall \boldsymbol{x},\boldsymbol{y} \in \mathcal{X}, | k(\boldsymbol{x},\boldsymbol{y})|<1$, the $\mathcal{I}_{n}[v](\boldsymbol{x}) < \infty$. 

The discrete approximation of the $n$-order kernel integral over  $N^n$  samples (with $N$ samples for each integral variable $\boldsymbol{y}_m$) is given by:
\begin{equation}
\begin{aligned}
\hat{\mathcal{I}}_{n}[v](\boldsymbol{x})
= \frac{1}{N^{n}} \sum_{i_1=1}^{N} \cdots \sum_{i_n=1}^{N}
&\left[ \prod_{m=1}^{n-1}
k(\boldsymbol{y}^{i_m}_m,
\boldsymbol{y}_{m+1}^{i_{m+1}}) \right] \\
&\times k(\boldsymbol{x}, \boldsymbol{y}_1^{i_1})
v(\boldsymbol{y}_n^{i_n})
\end{aligned}
\end{equation}
where $\boldsymbol{y}^{i_m}_m$ denotes the $i_m$-th sample of the $m$-th integral variable $\boldsymbol{y}_m$.

With the $n$-order kernel integral operator $\mathcal{I}_{n}[v](\cdot)$,  we propose the infinite-order kernel integral operator as Eq.~(\ref{InfiniteOp})
\begin{align}
\label{InfiniteOp}
    \mathcal{I}_{\infty}[v](\cdot) := \mathcal{I}_{1}[v](\cdot) + \alpha\mathcal{I}_{2}[v](\cdot) + \alpha^{2}\mathcal{I}_{3}[v](\cdot) + \cdots
\end{align}
with proper $\alpha$ to ensure series convergence. 

\subsubsection{Closed-form finite approximation}
Defining feature maps over $N$ samples as  $\boldsymbol{V}(Y) := [v(\boldsymbol{y}^{1}), \dots, v(\boldsymbol{y}^{N})]^{\top} \in \mathbb{R}^{N \times h}$ and the kernel vector $\boldsymbol{k}(\boldsymbol{x}) := \frac{1}{N}[k(\boldsymbol{x},\boldsymbol{y}^{1}), \dots, k(\boldsymbol{x},\boldsymbol{y}^{N})]^{\top} \in \mathbb{R}^{N \times 1}$, we  can formulate the discrete approximations of the $n^{th}$-order kernel integral $\mathcal{I}_{n}[v](\boldsymbol{x})$  in a matrix form:
\begin{align}
    \hat{\mathcal{I}}_{n}[v](x) = \boldsymbol{k}^{\top}(x)\boldsymbol{K}^{n\!-\!1}\boldsymbol{V}(Y)
\end{align}
where $\boldsymbol{K}$ denotes the kernel Gram matrix $\boldsymbol{K}(Y,Y)$ with its $ij^{th}$-element as $K_{ij} = \frac{1}{N} k(\boldsymbol{y}^{i}, \boldsymbol{y}^{j})$, and $\boldsymbol{K}^{n\!-\!1}$ denotes $({n\!-\!1})$ order of $\boldsymbol{K}$,  i.e.,     $\boldsymbol{K}^{n\!-\!1} = \underbrace{\boldsymbol{K} \boldsymbol{K} \cdots \boldsymbol{K} } _ {n-1 \; \text{times}}$, and $\boldsymbol{K}^0 = \boldsymbol{I}$. 

We provide the finite approximation of the proposed infinite-order kernel integral operator  as follows 
\begin{align}
     \hat{\mathcal{I}}_{\infty} [{v}]  (\boldsymbol{x}) & :=  \hat{\mathcal{I}}_1 [{v}]  (\boldsymbol{x}) + \alpha\hat{\mathcal{I}}_2 [{v}]  (\boldsymbol{x}) + \alpha^2 \hat{\mathcal{I}}_3 [{v}]  (\boldsymbol{x}) + \cdots 
\end{align}
Substituting matrix forms yields the geometric series:
\begin{equation}
\label{Fapp}
\begin{aligned}
\hat{\mathcal{I}}_{\infty}[v](\boldsymbol{x}) &= \boldsymbol{k}^{\top}(\boldsymbol{x})(\boldsymbol{I} + \alpha \boldsymbol{K} + \alpha^{2}\boldsymbol{K}^{2} + \cdots)\boldsymbol{V}(Y)
\end{aligned}
\end{equation}
When $\rho(\alpha \boldsymbol{K})<1$, where $\rho(\cdot)$ denotes the spectral radius, the positive-power Neumann series in Eq.~(\ref{Fapp}) converges and gives the closed-form resolvent:
\begin{align}
\label{CLFapp}
   \hat{\mathcal{I}}_{\infty}[v](\boldsymbol{x}) &=\boldsymbol{k}^{\top}(x)(\boldsymbol{I} - \alpha \boldsymbol{K})^{-1}\boldsymbol{V}(Y)
\end{align}
The same resolvent also admits a convergent inverse-power expansion when $\alpha\boldsymbol{K}$ is invertible and $\rho((\alpha\boldsymbol{K})^{-1})<1$:
\begin{align}
\label{InvPower}
    (\boldsymbol{I} - \alpha\boldsymbol{K})^{-1}
    &= -(\alpha\boldsymbol{K})^{-1}
    \left(\boldsymbol{I} - (\alpha\boldsymbol{K})^{-1}\right)^{-1} \\
    &= -\sum _{n=1}^{\infty}(\alpha\boldsymbol{K})^{-n}.
\end{align}
For a strictly positive-definite kernel evaluated on distinct sample points, $\boldsymbol{K}$ is symmetric positive definite and hence invertible; in this case the inverse-power series converges whenever the spectral norm of matrix $(\alpha \boldsymbol{K})^{-1}$ is less than one, i.e., $\|(\alpha \boldsymbol{K})^{-1} \|_2 = \frac{1}{|\alpha|  } \lambda_{max}(\boldsymbol{K}^{-1})= \frac{1}{|\alpha| \lambda_{min}(\boldsymbol{K})} <1$. 
 Equivalently $|\alpha|\lambda_{\min}(\boldsymbol{K})>1$.

\textbf{Remark:} The parameter $\alpha$ can either be positive or negative. By choosing $\alpha<0$, the operator $\boldsymbol{I}-\alpha\boldsymbol{K}$ avoids ill-posed problems and makes the training process significantly more stable.




\subsubsection{Two infinite-order neural operator constructions}
\label{TwoImpl}

While the closed-form approximation in Eq.~(\ref{CLFapp}) is derived in the matrix form on a single set of $N$ samples, extending the infinite-order idea to a $d$-dimensional product grid admits two natural neural operator constructions. We adopt the standard dimension-wise product base kernel,
\begin{align}
\label{ProdKernel}
    k(\boldsymbol{x},\boldsymbol{y}) = k_1(x_1,y_1) \times k_2(x_2,y_2) \times \cdots \times k_d(x_d,y_d),
\end{align}
which induces a kernel Gram matrix $\boldsymbol{K} \in \mathbb{R}^{N^d \times N^d}$ over the product grid that factorizes as the Kronecker product
\begin{align}
\label{KronK}
   \boldsymbol{K}  = \boldsymbol{K}_1 \otimes \boldsymbol{K}_2 \otimes \cdots \otimes \boldsymbol{K}_d ,
\end{align}
where $\boldsymbol{K}_j \in \mathbb{R}^{N \times N}$ is the one-dimensional kernel matrix on axis $j$ with elements $(\boldsymbol{K}_j)_{pq} = k_j(y_j^{p}, y_j^{q})$, and $y_j^{p}, y_j^{q}$ denote the $p^{th}$ and $q^{th}$ grid points in the $j$-th dimension. Based on this factorization, we define two alternative discrete infinite-order operators. They should be viewed as two distinct infinite-order neural operator constructions.

\textbf{IKNO-Vanilla.} The first construction applies Eq.~(\ref{CLFapp}) directly to the full kernel matrix on the product grid:
\begin{align}
\label{IKNOVanilla}
    \mathcal{K}_{\infty}^{\textnormal{Van}} \;:=\; (\boldsymbol{I}_M - \alpha \boldsymbol{K})^{-1},
    \quad M = N^d.
\end{align}
This is a faithful discretization of the infinite-order geometric series in Eq.~(\ref{Fapp}) on the $d$-dimensional grid: every cross-dimensional coupling term $\boldsymbol{K}^n$ is retained, so $\mathcal{K}_{\infty}^{\textnormal{Van}}$ inherits the full expressive power of the continuous operator $\mathcal{I}_{\infty}$.

\textbf{IKNO-TP.} The second construction is motivated by the equivalence between tensor products in function spaces and Kronecker products in matrix spaces. For Hilbert spaces $\mathcal{H}_j = L^2(\mathcal{X}_j)$ ($j=1,\dots,d$), the tensor product space satisfies
$\mathcal{H}_1 \otimes \cdots \otimes \mathcal{H}_d \;\cong\; L^2(\mathcal{X}_1 \times \cdots \times \mathcal{X}_d)$
and linear operators compose via
\begin{align}
    &(A_1 \otimes \cdots \otimes A_d)
    (f_1 \otimes \cdots \otimes f_d) \nonumber\\
    &\qquad = (A_1 f_1) \otimes \cdots \otimes (A_d f_d).
\end{align}
Interpreting each one-dimensional resolvent $(\boldsymbol{I}_N - \alpha \boldsymbol{K}_j)^{-1}$ as the infinite-order kernel operator acting on its own axis, we \emph{define} the multi-dimensional operator as the tensor product of these per-axis operators:
\begin{align}
\label{IKNOTP}
    \mathcal{K}_{\infty}^{\textnormal{TP}} \;:=\;&
    (\boldsymbol{I}_N - \alpha \boldsymbol{K}_1)^{-1}
    \otimes (\boldsymbol{I}_N - \alpha \boldsymbol{K}_2)^{-1} \nonumber\\
    &\otimes \cdots \otimes
    (\boldsymbol{I}_N - \alpha \boldsymbol{K}_d)^{-1} .
\end{align}
We emphasize that $\mathcal{K}_{\infty}^{\textnormal{TP}}$ is \emph{not} algebraically identical to $\mathcal{K}_{\infty}^{\textnormal{Van}}$; rather, it is an alternative operator defined by tensor composition of per-axis resolvents. Geometrically, $\mathcal{K}_{\infty}^{\textnormal{TP}}$ lifts the input into a structured latent space spanned by the tensor-product basis induced by the per-dimension kernels. This structured hypothesis class implicitly constrains the functional form of the learned operator and introduces a form of \emph{functional regularization}, which can be beneficial on problems with strong multi-scale or high-frequency components.

We refer to the neural operators built with $\mathcal{K}_{\infty}^{\textnormal{Van}}$ and $\mathcal{K}_{\infty}^{\textnormal{TP}}$ as \textbf{IKNO-Vanilla} and \textbf{IKNO-TP}, respectively. They share an identical end-to-end architecture (Sec.~\ref{sec:IKNO}) and differ only in the choice of discrete infinite-order operator.

\subsubsection{Fast computation}
\label{FastComp}

Directly forming and inverting the $M \times M$ resolvent in either Eq.~(\ref{IKNOVanilla}) or Eq.~(\ref{IKNOTP}) would incur $O(M^3) = O(N^{3d})$ operations, which is prohibitive even for moderate $N$ and $d$. For both variants, the Kronecker structure admits a scheme with $O(d \cdot N^{3})$ preprocessing and $O(d \cdot N \cdot M)$ cost per forward application, without ever materializing the $M \times M$ operator.

\textbf{Fast computation for IKNO-Vanilla.} For the full-kernel resolvent in Eq.~(\ref{IKNOVanilla}), we diagonalize each one-dimensional kernel matrix via a symmetric eigendecomposition $\boldsymbol{K}_j = \boldsymbol{U}_j \boldsymbol{\Lambda}_j \boldsymbol{U}_j^{\top}$. Using the Kronecker product property $(A_1\otimes\cdots\otimes A_d)(B_1\otimes\cdots\otimes B_d) = (A_1 B_1)\otimes\cdots\otimes(A_d B_d)$, the full resolvent admits the closed form
\begin{align}
\label{VanEig}
&(\boldsymbol{I}_M \!-\! \alpha\,
\boldsymbol{K}_1 \!\otimes \! \cdots \! \otimes \!
\boldsymbol{K}_d)^{-1} \nonumber\\
&\quad = (\boldsymbol{U}_1  \! \otimes \! \cdots  \! \otimes \!
\boldsymbol{U}_d) \,
\big[\boldsymbol{I}_M  \! - \! \alpha\,
\boldsymbol{\Lambda}_1  \! \otimes \! \cdots \! \otimes  \!
\boldsymbol{\Lambda}_d\big]^{-1} \nonumber\\
&\qquad \times
(\boldsymbol{U}_1 \! \otimes \! \cdots \! \otimes \!
\boldsymbol{U}_d)^{\top} ,
\end{align}
where the middle term is diagonal and given by
\begin{align}
\label{VanDiag}
&\big[\boldsymbol{I}_M - \alpha\,
\boldsymbol{\Lambda}_1 \otimes \cdots \otimes
\boldsymbol{\Lambda}_d\big]^{-1} \nonumber\\
&\quad = \operatorname{diag}\!\left(
\frac{1}{1 - \alpha\, \lambda^{(1)}_{i_1}
\lambda^{(2)}_{i_2} \cdots \lambda^{(d)}_{i_d}}
\right)_{(i_1,\dots,i_d)} ,
\end{align}
with $\lambda^{(j)}_{i_j}$ the $i_j$-th eigenvalue of $\boldsymbol{K}_j$. Preprocessing thus consists of $d$ eigendecompositions at $O(d \cdot N^{3})$ plus construction of the $M$ diagonal entries at $O(M)$. Each forward application applies $(\boldsymbol{U}_1\otimes\cdots\otimes\boldsymbol{U}_d)^{\top}$, the diagonal reweighting, and $(\boldsymbol{U}_1\otimes\cdots\otimes\boldsymbol{U}_d)$ through per-dimensional tensor-matrix multiplications at total cost $O(d \cdot N \cdot M)$.

\textbf{Fast computation for IKNO-TP.} By definition, $\mathcal{K}_{\infty}^{\textnormal{TP}}$ is already a Kronecker product of per-axis resolvents, so it reduces to inverting $d$ independent $N \times N$ matrices:
\begin{align}
\label{TPinv}
    \mathcal{K}_{\infty}^{\textnormal{TP}} \;=\; \bigotimes_{j=1}^{d} (\boldsymbol{I}_N - \alpha \boldsymbol{K}_j)^{-1} .
\end{align}
This yields a preprocessing cost of $O(d \cdot N^{3})$. Applying $\mathcal{K}_{\infty}^{\textnormal{TP}}$ to any feature tensor $\boldsymbol{V} \in \mathbb{R}^{M \times h}$ is then carried out as $d$ successive per-dimensional $N \times N$ matrix multiplications at total cost $O(d \cdot N \cdot M)$.

\textbf{Complexity comparison.} Both IKNO-Vanilla and IKNO-TP reduce the naive $O(N^{3d})$ matrix-inverse cost to $O(d \cdot N^{3})$ at preprocessing and $O(d \cdot N \cdot M)$ per forward pass. For both variants, applying the operator to features is dominated by a similar sequence of tensor-matrix multiplications along each dimension. IKNO-Vanilla incurs only a minor additional overhead from the symmetric eigendecomposition relative to direct inversion. In practice, the wall-clock runtimes of the two variants are nearly indistinguishable on all tested grids, so the choice between them is driven entirely by accuracy considerations (Sec.~\ref{sec:results}).

\subsubsection{Learnable multi-scale base kernel}
\label{sec:base_kernel}
To capture both smooth global structures and sharp local gradients, we design a learnable base kernel for each dimension $j$. We introduce learnable parameters $c_j$ and $\beta_j$ and $\gamma_j$ that rescale the coordinate distance, effectively reconfiguring the metric space for Gaussian and Laplace components respectively:
\begin{equation}
\begin{aligned}
k_{j}(x_{j}, y_{j})
&= c_j \Big(
\exp\left(-|\beta_j(x_{j} - y_{j})|^{2}\right) \\
&\qquad + \exp\left(-|\gamma_j(x_{j} - y_{j})|\right)
\Big)
\end{aligned}
\end{equation}
where $x_j$ and $y_j$ denote the element in $j^{th}$ dimension of input vector $\boldsymbol{x}$ and $\boldsymbol{y}$, respectively. Since both Gaussian and Laplacian kernels are strictly positive-definite, their nonnegative linear combination remains strictly positive-definite when the coefficients are non-degenerate; in our parameterization, this corresponds to $c_j>0$ and nonzero distance scales. Therefore, for distinct grid points, the induced Gram matrix is real symmetric positive definite.

 By jointly learning $\{c_j, \beta_j, \gamma_j \}$ for $j \in \{1,\cdots, d\}$, the model reconfigures its receptive field for each dimension, providing a flexible representation for complex anisotropic operators.


\subsection{IKNO architecture}
\label{sec:IKNO}

IKNO follows an encoder--processor--decoder architecture on a structured latent grid. Given point-cloud observations, an MLP first tokenizes geometry-enhanced input features. The infinite-order kernel encoder maps the point tokens to the latent grid; a Transformer processor performs global mixing on this grid; and the infinite-order kernel decoder maps the processed latent representation back to arbitrary query coordinates. Both IKNO-Vanilla and IKNO-TP use this architecture and differ only in the discrete infinite-order operator defined in Sec.~\ref{TwoImpl}. Detailed tokenization, encoding, decoding, and multi-scale kernel-branch designs are provided in Appendix~\ref{app:ikno_architecture}.
\begin{align}
\boldsymbol{V}_G &= \mathcal{K}_{\infty}\boldsymbol{K}_{GP}\boldsymbol{V}_P, \\
\boldsymbol{V}'_G &= \mathrm{TransformerEncoder}(\boldsymbol{V}_G), \\
\hat{u}(\boldsymbol{x}) &= \mathrm{MLP}_{\mathrm{head}}\!\left(\boldsymbol{k}^{\top}(\boldsymbol{x})\mathcal{K}_{\infty}\boldsymbol{V}'_G\right).
\end{align}

\section{Experiments}

In this section, we evaluate the performance of IKNO across a diverse range of benchmarks, spanning from classical physical systems to complex industrial-grade aerodynamic simulations.

\subsection{Experimental setup}

\textbf{Datasets.} We consider three categories of datasets to comprehensively assess the generalization and scalability of the proposed IKNO model: \textbf{Time-Independent Benchmarks}, \textbf{Time-Dependent Benchmarks} and \textbf{Industrial-scale Benchmarks}. More details about the benchmarks can be found in the Appendix~\ref{Datasets}.

\textbf{Baselines and Protocol.} To ensure a fair and rigorous comparison, we strictly follow the experimental protocols established by the current state-of-the-art model, GAOT\cite{wen2025geometry}. We directly adopt the reported results from the GAOT study for all baseline models, including Transolver\cite{wu2024transolver}, GINO\cite{li2023geometry}, GNOT\cite{hao2023gnot} and RIGNO\cite{mousavi2025rigno}. All models are evaluated using the same training and testing splits and identical evaluation metrics.

\begin{table*}[t]
\caption{Comparison of median relative $L_1$ error ($\%$) on time-independent benchmarks. Bold and underline indicate the best and second-best results across all methods, respectively.}
\label{tab:time_independent_full}
\begin{center}
\begin{small}
\begin{sc}
\resizebox{1\textwidth}{!}{
\begin{tabular}{@{}lcccccccc@{}}
\toprule
& \multicolumn{2}{c}{\textbf{Ours}} & \multicolumn{6}{c}{\textbf{Baselines}} \\
\cmidrule(lr){2-3} \cmidrule(lr){4-9}
\textbf{Dataset} & \textbf{IKNO-Vanilla} & \textbf{IKNO-TP} & \textbf{GAOT} & \textbf{RIGNO-18} & \textbf{Transolver} & \textbf{GNOT} & \textbf{UPT} & \textbf{GINO} \\
\midrule
Poisson C-Sines & \underline{1.42} & \textbf{1.00} & 3.10 & 6.83 & 77.3 & 100 & 100 & 20.0 \\
Poisson Gauss   & \underline{0.34} & \textbf{0.26} & 0.83 & 2.26 & 2.02 & 88.9 & 48.4 & 7.57 \\
Elasticity      & \textbf{0.93} & \underline{0.96} & 1.34 & 4.31 & 4.92 & 10.4 & 12.6 & 4.38 \\
NACA0012        & \textbf{3.76}   & \underline{4.22} & 6.81 & 5.30 & 8.69 & 6.89 & 16.1 & 9.01 \\
NACA2412        & \underline{4.99}   & \textbf{4.23} & 6.66 & 6.72 & 8.51 & 8.82 & 17.9 & 9.39 \\
RAE2822         & 4.97 & \textbf{4.70} & 6.61 & 5.06 & \underline{4.82} & 7.15 & 16.1 & 8.61 \\
\bottomrule
\end{tabular}
}
\end{sc}
\end{small}
\end{center}
\end{table*}

\begin{table*}[t]
\caption{Comparison of median relative $L_1$ error ($\%$) on time-dependent datasets. Bold and underline indicate the best and second-best results across all methods, respectively.}
\label{tab:time_full}
\begin{center}
\begin{small}
\begin{sc}
\resizebox{1\textwidth}{!}{
\begin{tabular}{@{}lcccccccc@{}}
\toprule
& \multicolumn{2}{c}{\textbf{Ours}} & \multicolumn{6}{c}{\textbf{Baselines}} \\
\cmidrule(lr){2-3} \cmidrule(lr){4-9}
\textbf{Dataset} & \textbf{IKNO-Vanilla} & \textbf{IKNO-TP} & \textbf{GAOT} & \textbf{RIGNO-18} & \textbf{GeoFNO} & \textbf{FNO DSE} & \textbf{UPT} & \textbf{GINO} \\
\midrule
NS-Gauss    & \textbf{1.48}   & \underline{1.69} & 2.91 & 2.29 & 41.1 & 38.4 & 92.5 & 13.1 \\
NS-PwC      & \underline{0.83}   & \textbf{0.63} & 1.50 & 1.58 & 26.0 & 56.7 & 100  & 5.85 \\
NS-SL       & 1.31 & \textbf{1.12} & \underline{1.21} & 1.28 & 13.7 & 22.6 & 51.5 & 4.48 \\
NS-SVS      & \underline{0.38} & \textbf{0.34} & 0.46 & 0.56 & 9.75 & 26.0 & 4.2  & 1.19 \\
CE-Gauss    & 7.30 & \textbf{6.22}   & \underline{6.40} & 6.90 & 42.1 & 30.8 & 64.2 & 25.1 \\
CE-RP       & 7.93 & \underline{5.85} & 5.97 & \textbf{3.98} & 18.4 & 27.7 & 26.8 & 12.3 \\
Wave-Layer  & 5.84 & \textbf{3.17}   & \underline{5.78} & 6.77 & 11.1 & 28.3 & 19.6 & 19.2 \\
Wave-C-Sines & \underline{2.86} & \textbf{1.70}   & 4.65 & 5.35 & 13.1 & 5.52 & 12.7 & 5.82 \\
\bottomrule
\end{tabular}
}
\end{sc}
\end{small}
\end{center}
\vspace{-10pt}
\end{table*}

\subsection{Implementation details}

We provide the detailed architectural configuration of our proposed IKNO model. Almost all parameters are kept consistent across all experiments to demonstrate the robustness of the architecture, with the exception of the grid resolution which is adapted to the spatial dimensionality and complexity of specific datasets.  More details can be found in the Appendix~\ref{Implementation}.

\subsection{Experimental results}
\label{sec:results}

\textbf{Results on Non Industrial Datasets.}

The experimental results for the median relative $L_1$ error on time-independent benchmarks are summarized in Table~\ref{tab:time_independent_full}. We compare the two variants introduced in Sec.~\ref{TwoImpl}: IKNO-Vanilla, which applies the full-kernel resolvent on the product grid, and IKNO-TP, which composes the per-axis resolvents via tensor product. On these time-independent datasets, IKNO-TP shows a clear performance advantage, achieving the best result on four of the six benchmarks and the second-best result on the two benchmarks where IKNO-Vanilla performs best.

The two variants exhibit complementary strengths, with IKNO-TP providing the strongest overall performance across the non-industrial benchmarks. Across the 14 time-independent and time-dependent datasets in Tables~\ref{tab:time_independent_full} and~\ref{tab:time_full}, IKNO-TP achieves the best result on ten benchmarks and the second-best result on the remaining four. IKNO-Vanilla remains a competitive non-tensorized alternative and attains the best result on several cases, including \textit{Elasticity}, \textit{NACA0012}, and \textit{NS-Gauss}. This pattern suggests that the tensor-product construction often provides a more robust inductive bias, while the full-coupling construction can still be advantageous on selected datasets.

These comprehensive results empirically validate that the infinite-order kernel integral -- under either construction -- yields a highly expressive neural operator, and that the two variants together cover a broad spectrum of PDE behaviors.

\textbf{Industrial Dataset Scalability.}

\begin{figure*}[!t]
    \centering
    \captionsetup{font=scriptsize,skip=0pt}
    \includegraphics[width=1\textwidth,trim=0 90 0 7,clip]{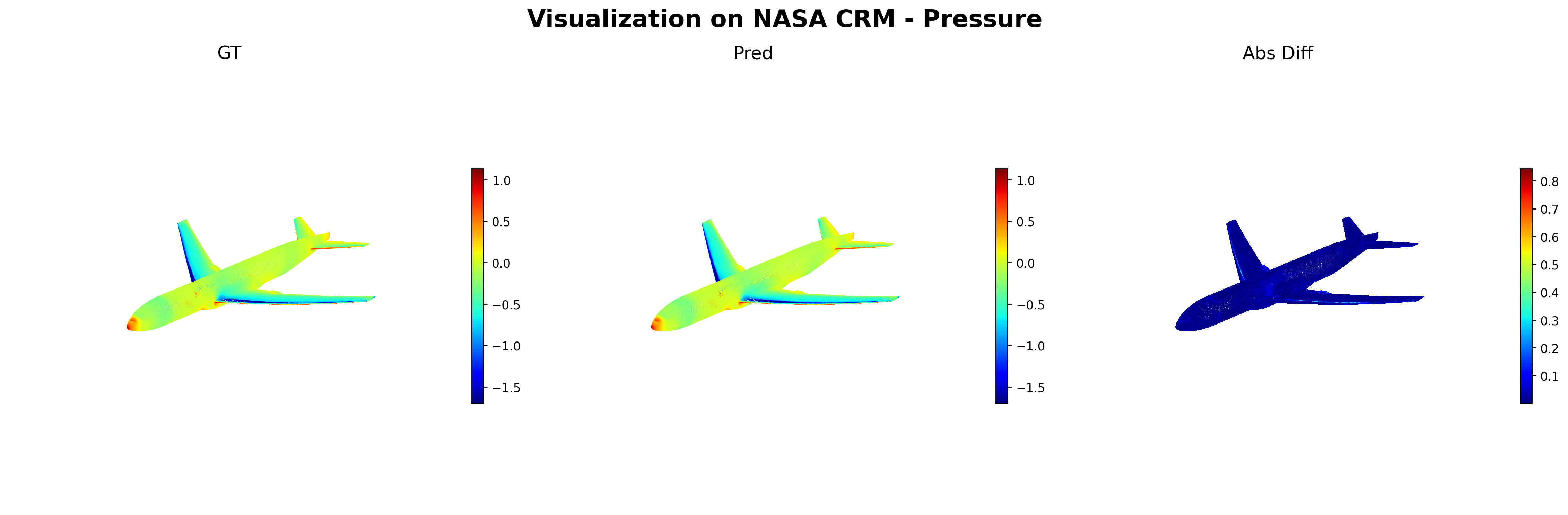}
    \caption{Visualization of IKNO predictions on the \textit{NASA CRM} industrial benchmark. The qualitative visualization further shows that IKNO produces coherent full-aircraft predictions on the high-resolution geometry, with visually small error patterns that indicate an accurate neural-operator approximation of the aerodynamic fields.}
    \label{fig:vis_nasa_crm}
\end{figure*}

To evaluate the robustness and scalability of IKNO in handling complex, industrial-grade aerodynamic simulations, we conduct experiments on the high-fidelity \textit{NASA CRM} benchmark~\cite{rivers2019nasa}. Unlike simplified 2D airfoils, this dataset involves 3D geometries with intricate topological features and high-Reynolds-number turbulent flows, posing a significant challenge for neural operators. Furthermore, each individual mesh in this dataset contains 440k points, presenting additional hurdles for GPU memory capacity and computational speed during model training.



We evaluate the two IKNO variants against the GAOT baseline on the large-scale industrial dataset \textit{NASA CRM}. The quantitative results, including Mean Squared Error and Mean Absolute Error for two groups of physical quantities (surface pressure $P$ and skin friction coefficient $C_f$), are summarized in Table~\ref{tab:large_scale_crm}. 


\begin{table}[h]
    \centering
    \caption{Comparison of GAOT and IKNO variants on the NASA CRM benchmark with MSE ($\times 10^{-2}$) and mean AE ($\times 10^{-1}$). Bold marks the best result per column. \textbf{P}: pressure. \textbf{Cf}: skin friction coefficient. \textbf{CRM}: NASA CRM dataset.}
\label{tab:large_scale_crm}
\resizebox{0.9\linewidth}{!}{
    \begin{tabular}{lcccc}
\toprule
\multirow{2}{*}{Model} & \multicolumn{2}{c}{P (CRM)} & \multicolumn{2}{c}{Cf (CRM)} \\
\cmidrule(r){2-3} \cmidrule(r){4-5}
& MSE & Mean AE & MSE & Mean AE \\
\midrule
GAOT          & 7.7170  & 1.6014 & 16.1091 & 2.2305 \\
IKNO-Vanilla  & 2.5108  & 0.8768 & 7.3943 & 1.4202 \\
IKNO-TP       & \textbf{2.1682} & \textbf{0.7801} & \textbf{6.0754} & \textbf{1.2372} \\
\bottomrule
\end{tabular}
}
\end{table}

All IKNO variants outperform GAOT by a substantial margin on every metric. IKNO-TP achieves the strongest performance, reducing the pressure MSE from $7.72$ to $2.17$ and the skin-friction MSE from $16.1$ to $6.08$, which demonstrates a remarkable performance improvement. IKNO-Vanilla also substantially improves over GAOT, showing that the full-coupling construction remains effective in this industrial regime. The further gain of IKNO-TP is particularly informative for \textit{NASA CRM}, which combines extremely high-resolution meshes (\(\sim\!440\mathrm{k}\) points per sample) with a very small training set (105 samples). Under such a low-data, high-dimensional setting, the structured tensor-product hypothesis class of IKNO-TP provides a more suitable representational bias and yields the best accuracy across both pressure and skin-friction fields.

\begin{table}[h]
    \centering
    \caption{Computational efficiency on \textit{Wave-C-Sines} (L40S GPU, identical training settings). ``IKNO (direct inverse)'' denotes the ablation in which the Kronecker-structured fast computation is disabled and $(\boldsymbol{I}_M - \alpha \boldsymbol{K})^{-1}$ is formed and inverted explicitly on the full product grid. Bold marks the best result per column among deployable methods (top three rows).}
\label{tab:efficiency}
\resizebox{0.9\linewidth}{!}{
    \begin{tabular}{lcc}
\toprule
\textbf{Model} & \textbf{Memory (GB)} & \textbf{Time / Epoch (s)} \\
\midrule
IKNO-TP                & \textbf{25.06} & \textbf{281}  \\
GAOT                   & 33.14          & 473           \\
Transolver             & 25.89          & 324           \\
IKNO (direct inverse)  & 24.84          & 1539          \\
\bottomrule
\end{tabular}
   }
\end{table}

\textbf{Computational efficiency.} Table~\ref{tab:efficiency} confirms that the Kronecker-structured fast computation preserves the practical scalability of IKNO: on \textit{Wave-C-Sines}, IKNO uses less memory and runs faster than GAOT and Transolver, while direct full-grid inversion causes a large runtime slowdown. Additional implementation details and analysis are provided in Appendix~\ref{app:efficiency}.

\textbf{Ablation summary.} Additional ablation studies in Appendix~\ref{app:ablation} show that the derivative prediction target is consistently the strongest temporal formulation, time-dependent datasets benefit substantially from increased latent grid resolution, and replacing the infinite-order kernel encoder with Cross-Attention severely degrades performance on challenging time-dependent benchmarks.

\section{Conclusions and future work}

In this paper, we have presented the Infinite-order Kernel Neural Operator (IKNO), a novel neural operator designed to address the expressive limitations of first-order kernel integral approximations in current neural operators. By introducing the infinite-order kernel integral to the encoder, IKNO achieves superior global information aggregation and propagation compared to traditional methods. We further introduce two complementary multi-dimensional infinite-order neural operator constructions, IKNO-Vanilla and IKNO-TP, together with Kronecker-structured fast computation schemes that make both constructions scalable on high-resolution latent grids.

Our empirical evaluation across 15 benchmarks demonstrates that IKNO achieves the best or highly competitive accuracy on nearly all time-independent and highly time-dependent PDE systems. Across these benchmarks, IKNO-TP is consistently the more robust variant, with particularly clear advantages on multi-scale, high-frequency, and low-data high-resolution industrial benchmarks. On the low-data, high-resolution \textit{NASA CRM} benchmark, both IKNO variants substantially improve over GAOT, with IKNO-TP achieving the strongest accuracy across all pressure and skin-friction metrics. These results also reveal practical limitations: IKNO depends on latent grid resolution, representation capacity can vary with the chosen infinite-order construction, and the initial kernel scales still require validation-based search rather than being automatically optimized to their best setting. Taken together, these findings position IKNO, especially IKNO-TP, as a scalable operator-learning framework for high-fidelity physics simulations.


\section*{Impact Statement}
This paper presents work whose goal is to advance scientific machine learning and neural operator methods for efficient PDE surrogate modeling. Potential societal impacts are primarily tied to the downstream scientific and engineering domains in which such models are deployed; we do not identify additional ethical concerns that must be specifically highlighted beyond standard model validation, reliability, and responsible use requirements for high-stakes simulations.

\bibliography{example_paper}

@article{wen2025geometry,
  title={Geometry aware operator transformer as an efficient and accurate neural surrogate for pdes on arbitrary domains},
  author={Wen, Shizheng and Kumbhat, Arsh and Lingsch, Levi and Mousavi, Sepehr and Zhao, Yizhou and Chandrashekar, Praveen and Mishra, Siddhartha},
  journal={arXiv preprint arXiv:2505.18781},
  year={2025}
}

@book{evans2022partial,
  title={Partial differential equations},
  author={Evans, Lawrence C},
  volume={19},
  year={2022},
  publisher={American mathematical society}
}

@book{vallis2017atmospheric,
  title={Atmospheric and oceanic fluid dynamics},
  author={Vallis, Geoffrey K},
  year={2017},
  publisher={Cambridge University Press}
}

@article{black1973pricing,
  title={The pricing of options and corporate liabilities},
  author={Black, Fischer and Scholes, Myron},
  journal={Journal of political economy},
  volume={81},
  number={3},
  pages={637--654},
  year={1973},
  publisher={The University of Chicago Press}
}

@book{murray2007mathematical,
  title={Mathematical biology: I. An introduction},
  author={Murray, James D},
  volume={17},
  year={2007},
  publisher={Springer Science \& Business Media}
}

@book{thomas2013numerical,
  title={Numerical partial differential equations: finite difference methods},
  author={Thomas, James William},
  volume={22},
  year={2013},
  publisher={Springer Science \& Business Media}
}

@book{hughes2003finite,
  title={The finite element method: linear static and dynamic finite element analysis},
  author={Hughes, Thomas JR},
  year={2003},
  publisher={Courier Corporation}
}

@article{li2020fourier,
  title={Fourier neural operator for parametric partial differential equations},
  author={Li, Zongyi and Kovachki, Nikola and Azizzadenesheli, Kamyar and Liu, Burigede and Bhattacharya, Kaushik and Stuart, Andrew and Anandkumar, Anima},
  journal={arXiv preprint arXiv:2010.08895},
  year={2020}
}

@article{li2020neural,
  title={Neural operator: Graph kernel network for partial differential equations},
  author={Li, Zongyi and Kovachki, Nikola and Azizzadenesheli, Kamyar and Liu, Burigede and Bhattacharya, Kaushik and Stuart, Andrew and Anandkumar, Anima},
  journal={arXiv preprint arXiv:2003.03485},
  year={2020}
}

@article{brandstetter2022message,
  title={Message passing neural PDE solvers},
  author={Brandstetter, Johannes and Worrall, Daniel and Welling, Max},
  journal={arXiv preprint arXiv:2202.03376},
  year={2022}
}

@article{li2023geometry,
  title={Geometry-informed neural operator for large-scale 3d pdes},
  author={Li, Zongyi and Kovachki, Nikola and Choy, Chris and Li, Boyi and Kossaifi, Jean and Otta, Shourya and Nabian, Mohammad Amin and Stadler, Maximilian and Hundt, Christian and Azizzadenesheli, Kamyar and others},
  journal={Advances in Neural Information Processing Systems},
  volume={36},
  pages={35836--35854},
  year={2023}
}

@article{rahman2022u,
  title={U-no: U-shaped neural operators},
  author={Rahman, Md Ashiqur and Ross, Zachary E and Azizzadenesheli, Kamyar},
  journal={arXiv preprint arXiv:2204.11127},
  year={2022}
}

@article{wu2024transolver,
  title={Transolver: A fast transformer solver for pdes on general geometries},
  author={Wu, Haixu and Luo, Huakun and Wang, Haowen and Wang, Jianmin and Long, Mingsheng},
  journal={arXiv preprint arXiv:2402.02366},
  year={2024}
}

@article{luo2025transolver++,
  title={Transolver++: An Accurate Neural Solver for PDEs on Million-Scale Geometries},
  author={Luo, Huakun and Wu, Haixu and Zhou, Hang and Xing, Lanxiang and Di, Yichen and Wang, Jianmin and Long, Mingsheng},
  journal={arXiv preprint arXiv:2502.02414},
  year={2025}
}

@article{wang2024latent,
  title={Latent neural operator for solving forward and inverse pde problems},
  author={Wang, Tian and Wang, Chuang},
  journal={Advances in Neural Information Processing Systems},
  volume={37},
  pages={33085--33107},
  year={2024}
}

@article{li2023fourier,
  title={Fourier neural operator with learned deformations for pdes on general geometries},
  author={Li, Zongyi and Huang, Daniel Zhengyu and Liu, Burigede and Anandkumar, Anima},
  journal={Journal of Machine Learning Research},
  volume={24},
  number={388},
  pages={1--26},
  year={2023}
}

@article{li2025geometric,
  title={Geometric operator learning with optimal transport},
  author={Li, Xinyi and Li, Zongyi and Kovachki, Nikola and Anandkumar, Anima},
  journal={arXiv preprint arXiv:2507.20065},
  year={2025}
}

@article{dosovitskiy2020image,
  title={An image is worth 16x16 words: Transformers for image recognition at scale},
  author={Dosovitskiy, Alexey},
  journal={arXiv preprint arXiv:2010.11929},
  year={2020}
}

@inproceedings{park2019deepsdf,
  title={Deepsdf: Learning continuous signed distance functions for shape representation},
  author={Park, Jeong Joon and Florence, Peter and Straub, Julian and Newcombe, Richard and Lovegrove, Steven},
  booktitle={Proceedings of the IEEE/CVF conference on computer vision and pattern recognition},
  pages={165--174},
  year={2019}
}

@article{vaswani2017attention,
  title={Attention is all you need},
  author={Vaswani, Ashish and Shazeer, Noam and Parmar, Niki and Uszkoreit, Jakob and Jones, Llion and Gomez, Aidan N and Kaiser, {\L}ukasz and Polosukhin, Illia},
  journal={Advances in neural information processing systems},
  volume={30},
  year={2017}
}

@article{rumelhart1986learning,
  title={Learning representations by back-propagating errors},
  author={Rumelhart, David E and Hinton, Geoffrey E and Williams, Ronald J},
  journal={nature},
  volume={323},
  number={6088},
  pages={533--536},
  year={1986},
  publisher={Nature Publishing Group UK London}
}

@article{li2022transformer,
  title={Transformer for partial differential equations' operator learning},
  author={Li, Zijie and Meidani, Kazem and Farimani, Amir Barati},
  journal={arXiv preprint arXiv:2205.13671},
  year={2022}
}

@inproceedings{hao2023gnot,
  title={Gnot: A general neural operator transformer for operator learning},
  author={Hao, Zhongkai and Wang, Zhengyi and Su, Hang and Ying, Chengyang and Dong, Yinpeng and Liu, Songming and Cheng, Ze and Song, Jian and Zhu, Jun},
  booktitle={International Conference on Machine Learning},
  pages={12556--12569},
  year={2023},
  organization={PMLR}
}

@book{galerkin1968rods,
  title={Rods and plates: series in some questions of elastic equilibrium of rods and plates},
  author={Galerkin, Boris Grigor{'}evich},
  year={1968},
  publisher={National Technical Information Service Springfield, VA, USA}
}

@article{mousavi2025rigno,
  title={RIGNO: A Graph-based framework for robust and accurate operator learning for PDEs on arbitrary domains},
  author={Mousavi, Sepehr and Wen, Shizheng and Lingsch, Levi and Herde, Maximilian and Raoni{\'c}, Bogdan and Mishra, Siddhartha},
  journal={arXiv preprint arXiv:2501.19205},
  year={2025}
}

@article{mildenhall2021nerf,
  title={Nerf: Representing scenes as neural radiance fields for view synthesis},
  author={Mildenhall, Ben and Srinivasan, Pratul P and Tancik, Matthew and Barron, Jonathan T and Ramamoorthi, Ravi and Ng, Ren},
  journal={Communications of the ACM},
  volume={65},
  number={1},
  pages={99--106},
  year={2021},
  publisher={ACM New York, NY, USA}
}

@inproceedings{rivers2019nasa,
  title={NASA common research model: a history and future plans},
  author={Rivers, Melissa B},
  booktitle={AIAA Aviation 2019 Forum},
  pages={3725},
  year={2019}
}

@article{loshchilov2017decoupled,
  title={Decoupled weight decay regularization},
  author={Loshchilov, Ilya and Hutter, Frank},
  journal={arXiv preprint arXiv:1711.05101},
  year={2017}
}

@inproceedings{pascanu2013difficulty,
  title={On the difficulty of training recurrent neural networks},
  author={Pascanu, Razvan and Mikolov, Tomas and Bengio, Yoshua},
  booktitle={International conference on machine learning},
  pages={1310--1318},
  year={2013},
  organization={Pmlr}
}

@article{altman2005standard,
  title={Standard deviations and standard errors},
  author={Altman, Douglas G and Bland, J Martin},
  journal={Bmj},
  volume={331},
  number={7521},
  pages={903},
  year={2005},
  publisher={British Medical Journal Publishing Group}
}

@article{herde2024poseidon,
  title={Poseidon: Efficient foundation models for pdes},
  author={Herde, Maximilian and Raonic, Bogdan and Rohner, Tobias and K{\"a}ppeli, Roger and Molinaro, Roberto and de B{\'e}zenac, Emmanuel and Mishra, Siddhartha},
  journal={Advances in Neural Information Processing Systems},
  volume={37},
  pages={72525--72624},
  year={2024}
}
\bibliographystyle{icml2026}

\newpage
\appendix
\onecolumn

\setcounter{table}{0} 
\renewcommand{\thetable}{\thesection.\arabic{table}}

\section{Problem formulation}
\label{app:problem_formulation}

We consider the problem of learning the neural operator of a Partial Differential Equation from data. Let $\Omega \subset \mathbb{R}^d$ be a bounded domain. We aim to learn a mapping between infinite-dimensional function spaces defined over $\Omega$ and with potentially a temporal domain $\Omega_T=[0, T]$.

\subsection{Time-independent case}
In the time-independent setting, the PDE describes a steady state where the solution $u$ depends on a given input condition $a$. Let $\mathcal{A}$ and $\mathcal{U}$ be two Banach spaces of functions. The time-independent PDE can be abstractly defined as:
\begin{equation}
\begin{cases}
\mathcal{L}(u(\boldsymbol{x}); a(\boldsymbol{x})) = 0, & \boldsymbol{x} \in \Omega \\
\mathcal{B}(u(\boldsymbol{x}); a(\boldsymbol{x})) = 0, & \boldsymbol{x} \in \partial \Omega
\end{cases}
\end{equation}
where $\mathcal{L}$ is a differential operator representing the corresponding PDE, $\mathcal{B}$ denotes the boundary operator, $a \in \mathcal{A}$ represents the input condition including coefficients, geometry parameters and source terms, and $u \in \mathcal{U}$ is the corresponding solution.

Our goal is to learn the solution operator $\mathcal{G}: \mathcal{A} \to \mathcal{U}$ such that:
\begin{equation}
u(\boldsymbol{x}) = \mathcal{G}(a)(\boldsymbol{x}), \quad \forall \boldsymbol{x} \in \Omega
\end{equation}
In our implementation, both $a$ and $u$ are observed on an arbitrary point cloud $\mathcal{P} = \{\boldsymbol{x}^i\}_{i=1}^{\bar{N}} \subset \Omega$. The input for each point $\boldsymbol{x}^i$ is a feature set $\{\boldsymbol{x}^i, a(\boldsymbol{x}^i)\}$.

\subsection{Time-dependent case}
For the time-dependent setting, we consider the evolution of the system over a time interval $\Omega_T=[0, T]$. Let $u(\boldsymbol{x}, t)$ denote the solution at position $\boldsymbol{x} \in \Omega$ and time $t \in \Omega_T$. The evolution is governed by the following PDE:
\begin{equation}
\begin{cases}
\frac{\partial u(\boldsymbol{x}, t)}{\partial t} = \mathcal{F}(u(\boldsymbol{x}, t); a(\boldsymbol{x}, t)), & (\boldsymbol{x}, t) \in \Omega \times \Omega_T \\
u(\boldsymbol{x}, 0) = u_0(\boldsymbol{x}), & \boldsymbol{x} \in \Omega
\end{cases}
\end{equation}
where $\mathcal{F}$ is a spatial differential operator representing the corresponding PDE, while the time-varying condition $a(\boldsymbol{x}, t)$ encapsulates all external forcing, source terms, and time-dependent parameters. $u_0(\boldsymbol{x})$ is the initial state of the system~\cite{evans2022partial}.

In this scenario, we aim to learn a trajectory operator $\mathcal{G}_{\text{td}}$ that maps the initial state and the entire history of conditions to the full solution trajectory:
\begin{equation}
u(\cdot, t)_{t \in \Omega_T} = \mathcal{G}_{\text{td}}(u_0, \{a(\cdot, t)\}_{t \in \Omega_T})
\end{equation}
The dataset consists of multiple trajectories sampled on arbitrary point clouds. For a single trajectory, given the discrete point set $\{\boldsymbol{x}^i\}_{i=1}^{\bar{N}}$ and time steps $\{t_j\}_{j=0}^{\bar{M}}$, the model is trained to predict $\{u(\boldsymbol{x}^i, t_j)\}$ for all $i, j$ based on the initial state $\{u_0(\boldsymbol{x}^i)\}$ and the sequence of conditions $\{a(\boldsymbol{x}^i, t_j)\}$.

For experiments on time-dependent benchmarks, we follow the unified lead-time prediction protocol of RIGNO~\cite{mousavi2025rigno}, which is also adopted by GAOT~\cite{wen2025geometry}. Details are provided in Appendix~\ref{app:temporal_framework}.

\section{Datasets}
\label{Datasets}

\textbf{Datasets.} We consider three categories of datasets to comprehensively assess the generalization and scalability of the proposed IKNO model:

\begin{itemize}
    \item \textbf{Time-Independent Benchmarks:} We select six time-independent datasets defined on complex, non-grid domains. These include \textit{Elasticity} from GeoFNO\cite{li2023fourier}, which involves stresses on an irregular mesh with a central hole; \textit{Poisson-Gauss} from RIGNO\cite{mousavi2025rigno}, solving the Poisson equation with Gaussian sources; and four datasets from GAOT\cite{wen2025geometry} —\textit{NACA0012}, \textit{NACA2412}, \textit{RAE2822}, and \textit{Poisson-C-Sines}. The former three are airfoil datasets represent challenging transonic flow conditions with significant geometric variations, while the \textit{Poisson-C-Sines} dataset features multi-scale solutions on a circular domain, testing the model's ability to handle unstructured inputs and high-frequency components.
    
    \item \textbf{Time-Dependent Benchmarks:} We choose eight time-dependent datasets introduced by RIGNO\cite{mousavi2025rigno}, covering the \textit{Compressible Euler equations}, \textit{incompressible Navier-Stokes equations}, and \textit{acoustic Wave equations}. These benchmarks involve complex physical phenomena such as shocks, sharp traveling waves, and diffraction. The non-linear interactions and reflections in these systems make them significantly harder to learn than time-independent cases.
    
    \item \textbf{Industrial-scale Benchmarks:} To evaluate the scalability of the IKNO model, we utilize the \textit{NASA CRM} (Common Research Model)\cite{rivers2019nasa}. This dataset serves as a rigorous industrial benchmark for civil transport aircraft, characterized by high-fidelity complexity: each individual sample contains approximately 440,000 spatial points. Despite this high dimensionality, the dataset provides a constrained set of only 105 training samples and 44 test samples. This large-scale data with small-scale samples nature poses a significant challenge for neural operators to generalize effectively. The model is required to predict pressure $P$ and skin friction coefficients $C_f$ based on aircraft geometry and simulation-specific parameters, such as the Mach number.
\end{itemize}

A detailed summary of these datasets is provided in Table~\ref{tab:dataset}.

\textbf{Dataset Licenses and Access.}
The RIGNO datasets used in this work are released on Zenodo under the Creative Commons Attribution 4.0 International (CC-BY-4.0) license. The GAOT datasets, Geo-FNO/Elasticity dataset, and NASA CRM dataset are publicly available through links provided by their original authors or maintainers, and we cite the corresponding original sources. However, we did not find explicit license statements for these datasets on their public access pages.


\begin{table}[h]
\centering
\caption{Summary of datasets used in this work.}
\label{tab:dataset}
\begin{tabular}{lll}
\toprule
\textbf{Abbreviation} & \textbf{Characteristic} & \textbf{Grid domain} \\
\midrule
Poisson-C-Sines      & Circular domain with sines              & F \\
Poisson-Gauss        & Gaussian source                          & T  \\
Elasticity           & Hole boundary distance                   & F \\
NACA0012             & Flow past NACA0012 airfoil               & F \\
NACA2412             & Flow past NACA2412 airfoil               & F \\
RAE2822              & Flow past RAE2822 airfoil                & F \\
NASA-CRM            & Surface pressure / friction coefficient  & F \\
\midrule
NS-Gauss             & Gaussian vorticity IC                    & T \\
NS-PwC               & Piecewise constant IC                    & T \\
NS-SL                & Shear layer IC                           & T \\
NS-SVS               & Sinusoidal vortex sheet IC               & T \\
CE-Gauss             & Gaussian vorticity IC                    & T \\
CE-RP                & 4-quadrant RP                            & T \\
Wave-Layer           & Layered wave medium                      & T \\
Wave-C-Sines         & Circular domain with sines               & F  \\
\bottomrule
\end{tabular}
\end{table}

\subsection{Preliminary finite-order kernel study}
\label{app:finite_order_motivation}

This appendix provides additional notation details for the preliminary study in Sec.~\ref{sec:motivation_finite_order}. The latent grid is denoted by $G=\{\boldsymbol{y}^j\}_{j=1}^{L^d}$, and the point-cloud input and query sets are denoted by $P=\{\boldsymbol{x}^i\}_{i=1}^{\bar{N}}$ and $Q=\{\boldsymbol{q}^r\}_{r=1}^{N_q}$, respectively. The matrix $\boldsymbol{K}_{GP}\in\mathbb{R}^{L^d\times\bar{N}}$ maps input points to latent grid points with entries $k(\boldsymbol{y}^j,\boldsymbol{x}^i)$, while $\boldsymbol{K}_{QG}\in\mathbb{R}^{N_q\times L^d}$ maps latent grid points to query locations with entries $k(\boldsymbol{q}^r,\boldsymbol{y}^j)$.

The token matrix over input points is $\boldsymbol{V}_{P}\in\mathbb{R}^{\bar{N}\times h}$. The matrices $\boldsymbol{V}_{G}$ and $\boldsymbol{V}'_{G}\in\mathbb{R}^{L^d\times h}$ denote latent-grid features before and after Transformer processing, respectively, and $\boldsymbol{V}_{Q}$ denotes the output tokens at query points before the final prediction head. All finite-order runs use a fixed linear-window kernel with radius $r=0.2$ and unit scale, together with a fixed propagation coefficient $\alpha=-0.15$; neither the kernel scale nor $\alpha$ is learned. The same architecture, training protocol, and dataset split are used across all finite orders; only the truncation order $p$ in $\boldsymbol{R}_p(G)$ is varied.

\section{Implementation details}
\label{Implementation}

We provide the detailed architectural configuration of our proposed IKNO model. Almost all parameters are kept consistent across all experiments to demonstrate the robustness of the architecture, with the exception of the grid resolution which is adapted to the spatial dimensionality and complexity of specific datasets.

\subsection{Detailed IKNO architecture}
\label{app:ikno_architecture}

\subsubsection{Architectural overview}
IKNO is designed to learn the solution operator of PDEs on complex, non-standard geometries. Unlike traditional operators that rely on local message passing or fixed-grid spectral methods, IKNO establishes a continuous mapping between unstructured point clouds and a latent grid through the infinite-order kernel integral. The architecture consists of four primary stages: (i) Feature Tokenization with geometry enhancement, (ii) Infinite Kernel Encoding from points to a latent grid, (iii) a Latent Processor for global feature mixing, and (iv) Infinite Kernel Decoding back to the physical quantities at query coordinates.

\subsubsection{Tokenization with geometry enhancement}
To ensure the model well captures the information of  the underlying manifold of the PDE domain, we employ a coordinate-based encoding strategy inspired by NeRF\cite{mildenhall2021nerf}. For an input point $\boldsymbol{x}^i \in \Omega$ with associated condition $a(\boldsymbol{x}^i)$, we apply a positional encoding feature map with $l=1$ frequency levels to capture fundamental spatial variations:
\begin{equation}
\psi(\boldsymbol{x}^i) = \left( \boldsymbol{x}^i, \cos(\boldsymbol{x}^i), \sin(\boldsymbol{x}^i) \right) \in \mathbb{R}^{3d}
\end{equation}
where $\text{cos}(\cdot)$ and $\text{sin}(\cdot)$ denotes element-wise operation.

The augmented features are concatenated and projected into a high-dimensional latent space via an embedding MLP to produce the initial tokens $v(\boldsymbol{x}^i)$:
\begin{equation}
v(\boldsymbol{x}^i) = \text{MLP}(\text{Concat}(\psi(\boldsymbol{x}^i), a(\boldsymbol{x}^i)))
\end{equation}

\subsubsection{Infinite kernel encoding}
The encoding stage maps the scattered point cloud tokens $\{v(\boldsymbol{x}^i)\}_{i=1}^{\bar{N}}$ onto a structured latent grid $G$ with grid points $\{\boldsymbol{y}^j\}_{j=1}^M$. To capture infinite-order global dependencies, we utilize the closed-form operator given in Eq.~\eqref{CLFapp}. The latent grid representation $\boldsymbol{V}_G \in \mathbb{R}^{M \times h}$ is generated as follows:
\begin{equation}
\boldsymbol{V}_G =  (\boldsymbol{I}_M - \alpha \boldsymbol{K})^{-1} \boldsymbol{K}_{GP}\boldsymbol{V}_P
\end{equation}
where $\boldsymbol{K}_{GP} \in \mathbb{R}^{M\times {\bar{N}}}$ denotes kernel matrix such that its  element in $j^{th}$-row and $i^{th}$-column is $k(\boldsymbol{y}^j,\boldsymbol{x}^i)$, and $\boldsymbol{V}_P = [v(\boldsymbol{x}^1),\cdots, v(\boldsymbol{x}^{\bar{N}})]^\top \in \mathbb{R}^{{\bar{N}} \times h}$ denotes the embedding feature matrix over input data.

By utilizing the infinite-order formulation, each grid point $\boldsymbol{y}^j$ adaptively aggregates information from all input data points, implicitly considering high-order physical correlations beyond simple linear kernels. In practice, multiple kernel results are fused through concatenation followed by an encoder fusion MLP to obtain the unified latent grid representation.

\subsubsection{Latent processor}
On the latent grid $G$, we employ a series of Transformer Encoder blocks to perform non-linear mixing of the grid features. This stage refines the global physical states in a latent space where the topology is regular and computationally efficient. Let $\boldsymbol{V}_G'$ denote the processed latent features:
\begin{equation}
\boldsymbol{V}_G' = \text{TransformerEncoder}(\boldsymbol{V}_G)
\end{equation}
$\boldsymbol{V}_G'$ is constructed by our TransformerEncoder.

\subsubsection{Infinite kernel decoding}
The decoding stage reconstructs the solution at arbitrary query coordinates $\boldsymbol{x}$ by projecting information from the latent grid $G$ back to the continuous domain. This process utilizes the same infinite-order kernel logic to maintain consistency in the functional representation:
\begin{equation}
v_{\text{out}}(\boldsymbol{x}) = \hat{\mathcal{I}}_{\infty}[v'](\boldsymbol{x}) = \boldsymbol{k}^{\top}(\boldsymbol{x}) (\boldsymbol{I}_M-\alpha \boldsymbol{K})^{-1} \boldsymbol{V}_G'
\end{equation}
where $\boldsymbol{V}_G' \in \mathbb{R}^{M \times h}$ denotes the embedding feature matrix over the latent grids.

Similar to the encoding stage, the output tokens will pass through an output head MLP to yield the predicted physical quantities $\hat{u}(\boldsymbol{x})$:
\begin{equation}
\hat{u}(\boldsymbol{x}) = \text{MLP}_{\text{head}}(v_{\text{out}}(\boldsymbol{x}))
\end{equation}

\subsubsection{Multi-scale kernel design}
To capture anisotropic physical gradients and various interaction scales, IKNO instantiates $Q=3$ independent kernel branches. Each branch $q \in \{1, \dots, Q\}$ utilizes the base kernel described in Sec.~\ref{sec:base_kernel}. Crucially, both the coefficients $c_{q,j}$ and the scaling parameters $\beta_{q,j}$, $\gamma_{q,j}$ are learnable for each branch $q$ and each spatial dimension $j$. The dimension-wise $c_{q,j}$ allows the model to adaptively control the propagation density of physical information along different axes, while $\beta_{q,j}$ and $\gamma_{q,j}$ reconfigure the metric space to capture both sharp local gradients and smooth global structures. The multi-scale features are concatenated and integrated using fusion MLPs during both the encoding and decoding phases, enabling a comprehensive representation of complex multiphysics interactions.

\subsection{Unified operator framework for temporal evolution}
\label{app:temporal_framework}

Following RIGNO~\cite{mousavi2025rigno}, and consistent with the GAOT~\cite{wen2025geometry} setting, we reformulate time-dependent trajectory prediction as a unified lead-time operator learning problem. This transition allows the neural operator to process time-dependent data as a collection of generalized snapshots, effectively treating the lead-time prediction as a time-independent mapping task.

Specifically, for a discrete trajectory sampled at $\{t_j\}_{j=0}^{\bar{M}}$, we consider any pair of time instances $(t_i, t_j)$ where $t_j = t_i + \tau$ and $\tau > 0$ denotes the lead time. To transform the evolution into a time-independent form, we augment the input condition $a(\boldsymbol{x})$ by incorporating the state at the current time $t_i$ and the temporal coordinates. The augmented input feature $\tilde{a}$ is defined as:
\begin{equation}
\tilde{a}(\boldsymbol{x}) \coloneqq \left\{ u(\boldsymbol{x}, t_i), a(\boldsymbol{x}, t_i), t_i, \tau \right\}, \quad \boldsymbol{x} \in \Omega
\end{equation}
where $u(\boldsymbol{x}, t_i)$ serves as the initial condition for the subsequent evolution over the interval $\tau$. Under this formulation, the trajectory operator $\mathcal{G}_{\text{td}}$ is simplified to a unified operator $\mathcal{G}: \tilde{\mathcal{A}} \to \mathcal{U}$, where $\tilde{\mathcal{A}}$ is the augmented input space.

We define three distinct prediction modes for the target quantity $y(\boldsymbol{x}) = \mathcal{G}(\tilde{a})(\boldsymbol{x})$:
\begin{itemize}
    \item \textbf{Direct Prediction:} The model directly maps the augmented input to the future state:
    \begin{equation}
        \mathcal{G}(\tilde{a})(\boldsymbol{x}) = u(\boldsymbol{x}, t_i + \tau)
    \end{equation}
    
    \item \textbf{Residual Prediction:} The model learns the incremental change between time steps:
    \begin{equation}
        \mathcal{G}(\tilde{a})(\boldsymbol{x}) = u(\boldsymbol{x}, t_i + \tau) - u(\boldsymbol{x}, t_i)
    \end{equation}
    
    \item \textbf{Derivative Prediction:} The model predicts the average rate of change over time:
    \begin{equation}
        \mathcal{G}(\tilde{a})(\boldsymbol{x}) = \frac{u(\boldsymbol{x}, t_i + \tau) - u(\boldsymbol{x}, t_i)}{\tau}
    \end{equation}
\end{itemize}

By adopting this unified framework, the neural operator is capable of handling irregularly sampled temporal data and zero-shot super-resolution in the time domain, utilizing the same architecture as the time-independent solver.

\subsection{Model parameter configuration}
For the latent space processor, we utilize a Transformer-based backbone with $d_{model}=256$, consisting of 12 layers and 8 attention heads. The feed-forward expansion factor is set to 4. For geometry enhancement, we employ a NeRF-inspired positional encoding with $l=1$ frequency levels, which includes the original input coordinates to preserve linear spatial relationships.

The multi-scale kernel mechanism consists of $Q=3$ independent branches. The decay coefficients $\alpha$ and scaling parameters are initialized as shown in Table~\ref{tab:params} and are optimized during training, allowing the model to adaptively learn the physical correlation scales.

\subsection{Model parameters}

\begin{table*}[h]
\centering
\caption{IKNO implementation parameters and default settings used in all experiments.}
\begin{tabular}{lll}
\toprule
\textbf{Parameter} & \textbf{Default Value} & \textbf{Description} \\
\midrule
\texttt{d\_model} & 256 & Transformer Hidden dimension \\
\texttt{num\_heads} & 8 & Transformer Attention heads \\
\texttt{num\_layers} & 12 & Transformer layers \\
\texttt{ff\_mult} & 4 & Transformer FFN expansion ratio \\
\texttt{dropout} & 0.0 & Transformer Dropout rate  \\
\texttt{nerf\_frequencies} & 1 & Positional encoding frequencies \\
\texttt{nerf\_include\_input} & True & Include raw coordinates \\
\texttt{nerf\_log\_sampling} & True & Log-spaced frequencies \\
\texttt{nerf\_scale} & 1.0 & Positional encoding scale \\
\texttt{grid\_size} & Task-dependent & Latent grid resolution \\
\texttt{grid\_min} & -1.0 & Latent grid lower bound \\
\texttt{grid\_max} & 1.0 & Latent grid upper bound \\
\texttt{num\_kernels} & 3 & Number of kernels \\
\texttt{kernel\_alpha} & $[-1, -1, -1]$ & Initial decay coefficients \\
\texttt{kernel\_alpha\_grad} & True & Learn decay coefficients \\
\texttt{init\_scales\_l1} & Base $[1, 2, 4]$ & Laplacian kernel scale base \\
\texttt{init\_scales\_l2} & Base $[1, 2, 4]$ & Gaussian kernel scale base \\
\texttt{scales\_grad} & True & Learn kernel scales \\
\bottomrule
\end{tabular}
\label{tab:params}
\end{table*}

\subsection{Latent grid settings}
While the majority of model parameters remain fixed, we adjust the resolution of the latent grid $G$ to accommodate the varying complexity and dimensionality of the benchmarks:
\begin{itemize}
    \item \textbf{Time-independent Datasets:} We employ a 2D grid with a resolution of $[24, 24]$.
    \item \textbf{Time-dependent Datasets:} To capture temporal evolution, the grid resolution is increased to $[32, 32]$.
    \item \textbf{Industrial-scale Datasets:}For the \textit{NASA CRM} dataset, the grid resolution is set to $[24, 24, 24]$. 
\end{itemize}

\section{Training details}


\begin{table*}[t]
\centering
\caption{Training, validation, and test sizes of datasets used in this work.}
\label{tab:dataset_split}
\begin{tabular}{lccc}
\toprule
\textbf{Abbreviation} & \textbf{Train size} & \textbf{Val size} & \textbf{Test size} \\
\midrule
Poisson-C-Sines  & 2048 & 256 & 256 \\
Poisson-Gauss    & 2048 & 256 & 256 \\
Elasticity       & 1024 & 256 & 256 \\
NACA0012         & 5000 & 256 & 256 \\
NACA2412         & 5000 & 256 & 256 \\
RAE2822          & 5000 & 256 & 256 \\
NASA-CRM         & 105  & 0  & 44  \\
\midrule
NS-Gauss         & 1024 & 256 & 256 \\
NS-PwC           & 1024 & 256 & 256 \\
NS-SL            & 1024 & 256 & 256 \\
NS-SVS           & 1024 & 256 & 256 \\
CE-Gauss         & 1024 & 256 & 256 \\
CE-RP            & 1024 & 256 & 256 \\
Wave-Layer       & 1024 & 256 & 256 \\
Wave-C-Sines     & 1024 & 256 & 256 \\
\bottomrule
\end{tabular}
\end{table*}

\subsection{Loss function}
To train the neural operator within the unified framework, we utilize the \textbf{relative $L_2$ loss} as our optimization objective. Given a set of points $\mathcal{P} = \{\boldsymbol{x}^i\}_{i=1}^n$, let $\hat{y}(\boldsymbol{x}^i)$ be the model prediction and $y(\boldsymbol{x}^i)$ be the ground truth defined according to the prediction modes. The loss for a single sample is defined as:
\begin{equation}
\mathcal{L}(y, \hat{y}) = \frac{\sqrt{\sum_{i=1}^n \|y(\boldsymbol{x}^i) - \hat{y}(\boldsymbol{x}^i)\|_2^2}}{\sqrt{\sum_{i=1}^n \|y(\boldsymbol{x}^i)\|_2^2}}
\end{equation}
For time-dependent trajectories, the total loss is averaged over all sampled time pairs $(t_i, t_j)$ within the batch.

\subsection{Training parameters and hardware}
All models are optimized using the AdamW\cite{loshchilov2017decoupled} optimizer with a cosine annealing learning rate scheduler. We apply gradient clipping\cite{pascanu2013difficulty} to ensure training stability, especially for complex time-dependent systems. All experiments are conducted on a single GPU. The time-dependent benchmarks are trained on a single NVIDIA L40S GPU, the time-independent benchmarks are trained on a single NVIDIA A100 Tensor Core GPU, and the \textit{NASA CRM} benchmark is trained on a single NVIDIA H200 Tensor Core GPU.

\subsection{Data partitioning and preprocessing}
\textbf{Dataset Splitting.}
Following the experimental setting of GAOT~\cite{wen2025geometry}, we use the same dataset partitioning strategy for all shared benchmarks except \textit{NASA CRM}. Specifically, the first $N_{\text{train}}$ samples are used for training, the subsequent $N_{\text{val}}$ samples are reserved for validation and hyperparameter search, and the final $N_{\text{test}}$ samples, counted backward from the end of the dataset, are held out exclusively for final metric computation. Since the \textit{NASA CRM} benchmark does not provide a validation split, only on this dataset do we use the training loss for hyperparameter selection.

\textbf{Hyperparameter Search.}
We only search the initial kernel scale hyperparameter. Starting from the base scale vector $[1,2,4]$, we evaluate multiplicative factors in $\{1,2,4\}$, corresponding to candidate initial scale vectors $[1,2,4]$, $[2,4,8]$ and $[4,8,16]$. Each candidate is trained for $20\%$ of the total training epochs, after which we select the candidate with the lowest validation loss and continue training it for the remaining epochs. Since \textit{NASA CRM} has no validation split, we use the training loss for this selection step.

\textbf{Data Pre-processing and Sampling.} 
For datasets labeled as \textit{grid domain} in Table~\ref{tab:dataset_split}, we transform them into an \textit{arbitrary domain} format to test the discretization-invariance of our operator. We follow the point cloud sampling strategy proposed in GAOT \cite{wen2025geometry}. Specifically, since the datasets provided by RIGNO \cite{mousavi2025rigno} already contain shuffled physical coordinates, we directly sample the first 9,216 points for each instance to construct the arbitrary point cloud input.

\textbf{Normalization.} 
To stabilize the training process, we apply Z-score normalization\cite{altman2005standard} to both the input features and target outputs. For any scalar or vector field $v$, the normalized field $\hat{v}$ is computed as:
\begin{equation}
    \hat{v}(\boldsymbol{x}) = \frac{v(\boldsymbol{x}) - \mu_v}{\sigma_v + \epsilon}
\end{equation}
where $\mu_v$ and $\sigma_v$ are the mean and standard deviation calculated across the training set, respectively, and $\epsilon=10^{-10}$ is a small constant for numerical stability.

\textbf{Temporal Data Handling.} 
For time-dependent PDEs, we adopt the \textit{all2all} training strategy introduced in Poseidon\cite{herde2024poseidon}. For a given trajectory with time steps $\{t_0, t_1, \dots, t_m\}$, we generate all possible pairs $(t_i, t_j)$ where $t_j = t_i + \tau$ and $\tau > 0$. Each pair constitutes a single training sample with current time $t_i$ and lead time $\tau$ as additional input scalars. To ensure consistent feature scaling, these temporal coordinates ($t_i$ and $\tau$) also undergo independent Z-score normalization based on their distributions in the training set.

Furthermore, when employing the Residual or Derivative prediction modes, the target quantity $y(\boldsymbol{x})$ is fundamentally transformed from the original state distribution. Consequently, we recompute the normalization statistics  specifically for these Residual-based or Derivative-based targets. This ensures that the loss function remains well-conditioned and that the model's output range is appropriately scaled to the transformed label space.

\section{Statistical robustness}
\label{app:statistical_robustness}

To evaluate the sensitivity of IKNO-TP to random initialization and stochastic training effects, we repeat representative experiments on four datasets with five random seeds. Table~\ref{tab:statistical_robustness} reports the mean and sample standard deviation of the median relative $L_1$ error ($\%$).

\begin{table}[h]
\centering
\caption{Statistical robustness of IKNO-TP over five random seeds. We report the mean and sample standard deviation of the median relative $L_1$ error ($\%$).}
\label{tab:statistical_robustness}
\begin{small}
\begin{sc}
\begin{tabular}{lc}
\toprule
\textbf{Dataset} & \textbf{Mean $\pm$ Std.} \\
\midrule
Poisson-C-Sines & $1.01 \pm 0.02$ \\
Elasticity      & $0.97 \pm 0.02$ \\
Poisson-Gauss   & $0.26 \pm 0.01$ \\
NS-SL           & $1.12 \pm 0.03$ \\
\bottomrule
\end{tabular}
\end{sc}
\end{small}
\end{table}

The standard deviations are small across all four datasets, indicating that IKNO-TP is stable with respect to random seed variation and that the reported performance is not driven by a single favorable run.

\section{Additional ablation studies}
\label{app:ablation}

In this section, we conduct a series of ablation experiments to evaluate the impact of the training objectives, latent grid configurations, and the proposed infinite kernel mechanism on the performance of the IKNO model.

\subsection{Impact of temporal prediction modes}
For time-dependent datasets, we investigate three distinct training formulations: \textit{Direct}, \textit{Residual}, and \textit{Derivative} prediction. While previous studies in neural operators have explored these variations, we re-examine their efficacy within the IKNO framework. As summarized in Table~\ref{tab:temporal_prediction}, the \textbf{Derivative} mode consistently yields the lowest relative error across all benchmarks, whereas the other two modes exhibit varying degrees of performance. 

\begin{table}[ht]
\centering
\caption{Comparison of temporal prediction modes. We report the \textbf{median relative $L_1$ error ($\%$)} for two time-dependent datasets. The Derivative mode demonstrates the highest precision.}
\label{tab:temporal_prediction}
\begin{small}
\begin{sc}
\begin{tabular}{lccc}
\toprule
Dataset & Direct & Residual & Derivative \\
\midrule
Wave-Layer & 8.40 & 8.66 & \textbf{8.16} \\
CE-Gauss   & 9.77 & 9.73 & \textbf{9.52} \\
\bottomrule
\end{tabular}
\end{sc}
\end{small}
\end{table}

We hypothesize that the superiority of the derivative mode stems from the explicit normalization by the time interval $\tau$. By predicting $\frac{u(t+\tau) - u(t)}{\tau}$, the model effectively amplifies the learning signal for small $\tau$, preventing the gradients from vanishing and ensuring numerical stability during training. However, it is worth noting that the performance gap between these modes remains relatively narrow, suggesting that IKNO is robust to the choice of optimization target.

\subsection{Scalability of latent grid resolution}
\begin{figure}
    \centering
    \includegraphics[width=0.8\linewidth]{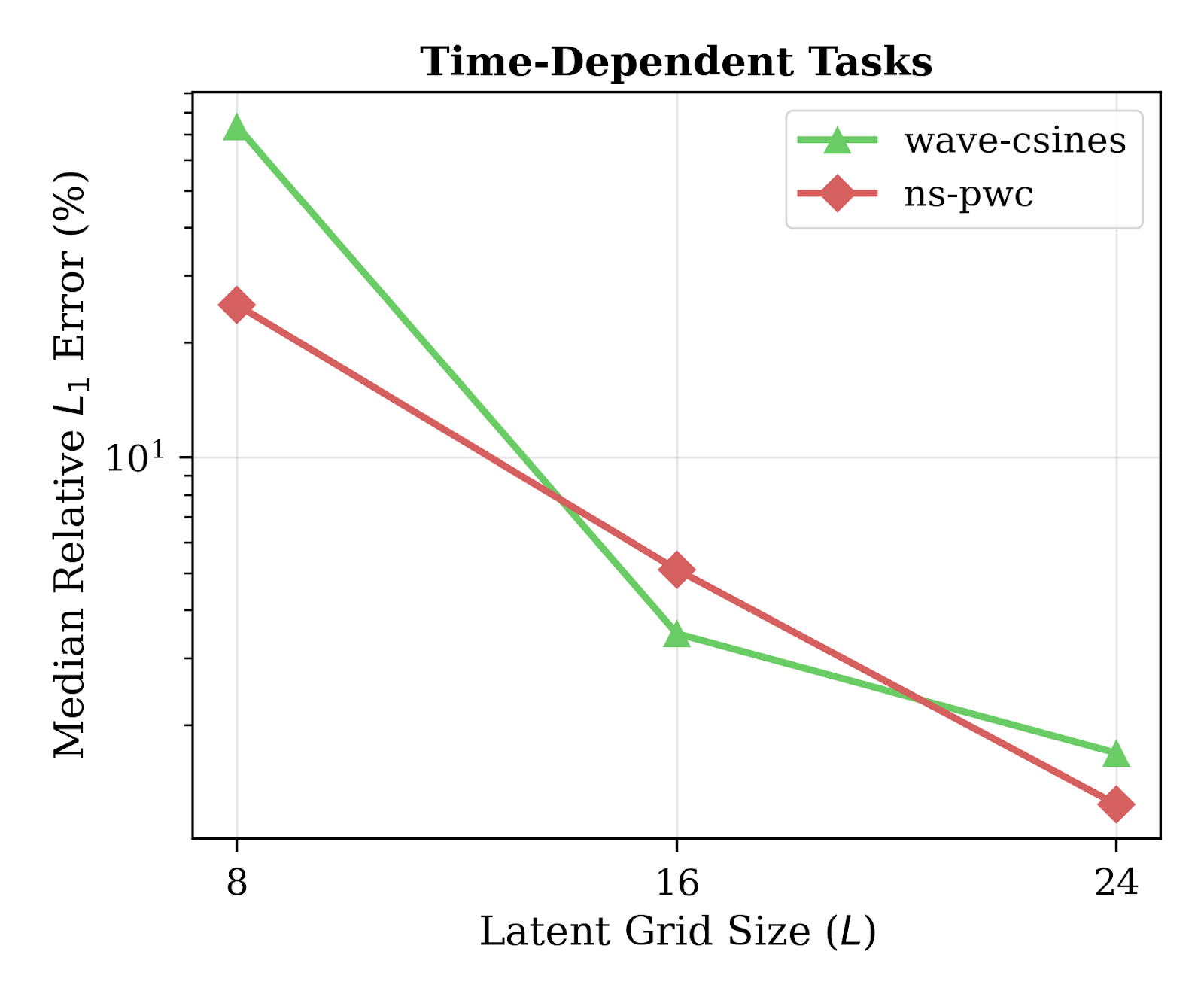}
    \caption{Ablation study on the latent grid size $L$ (per-axis resolution of the latent grid $G$) on time-dependent datasets. For instance, $L=32$ corresponds to a $32 \times 32$ latent grid.}
    \label{fig:g1}
\end{figure}

\begin{figure}
    \centering
    \includegraphics[width=0.8\linewidth]{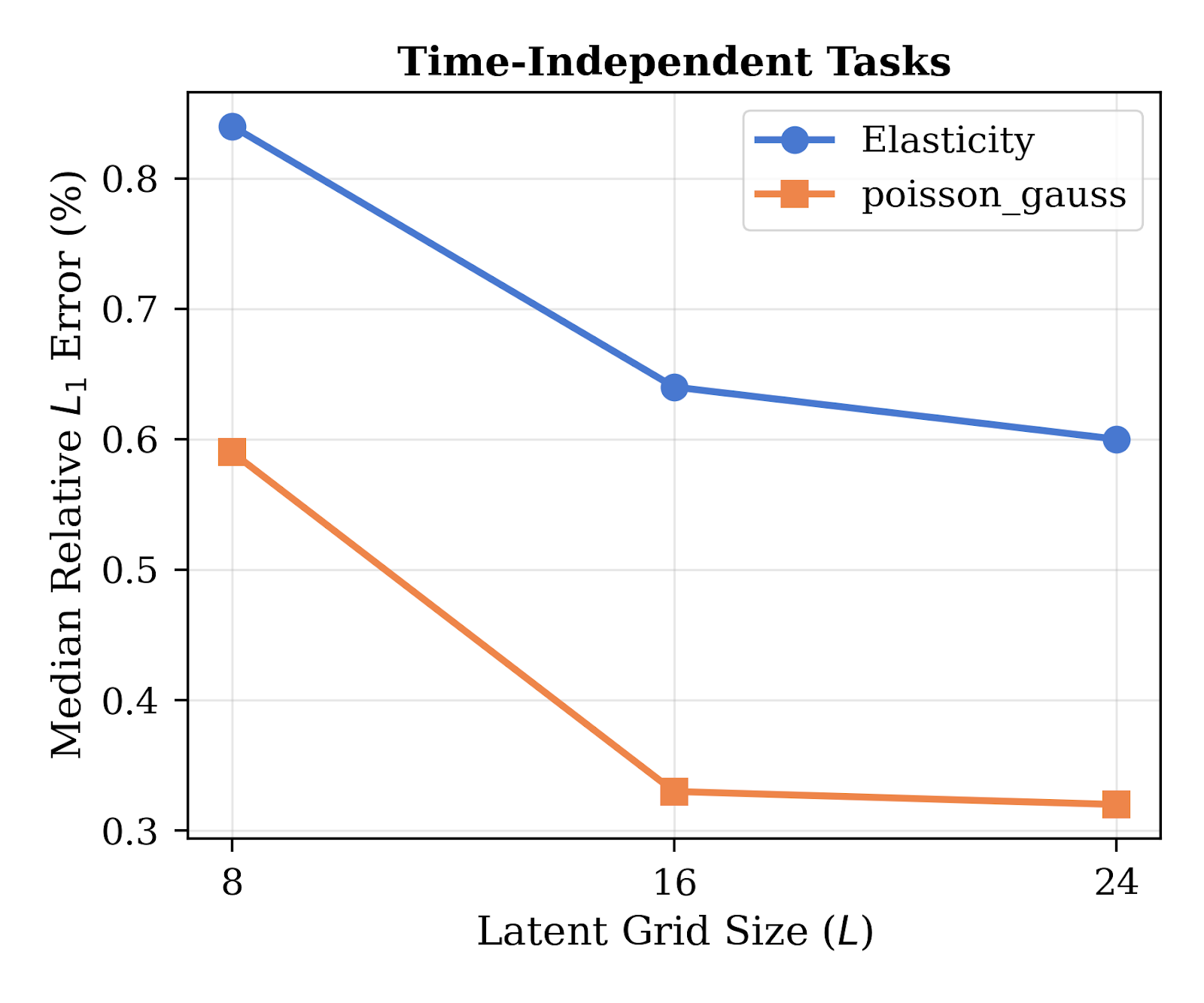}
    \caption{Ablation study on the latent grid size $L$ (per-axis resolution of the latent grid $G$) on time-independent datasets. For instance, $L=24$ corresponds to a $24 \times 24$ latent grid.}
    \label{fig:g2}
\end{figure}

We analyze the sensitivity of the model to the latent grid size $L$, defined as the per-axis resolution of the latent grid $G$, so that the grid contains $L^{d}$ points in total, where $d$ is the spatial dimension. Across four representative datasets, Figures~\ref{fig:g2} and~\ref{fig:g1} illustrate the performance trends for time-independent and time-dependent PDEs, respectively. In time-independent tasks, increasing the latent grid resolution leads to marginal performance gains. In contrast, for time-dependent systems, the model exhibits a critical dependence on the latent capacity. At low resolutions, the model fails to converge or produces high errors, whereas increasing the grid size results in a dramatic improvement in accuracy. This suggests that time-dependent trajectories contain higher-frequency components that necessitate a denser latent representation to be captured effectively.

\subsection{Necessity of the infinite kernel mechanism}
\label{app:ablation_kernel}

To verify the importance of the infinite-order kernel formulation, we compare our IKNO encoder against two degraded alternatives that ablate different aspects of it.

\textbf{First-order Kernel.} The first alternative replaces the infinite-order resolvent in the encoder with its $p$-th order truncation of the geometric series in Eq.~(\ref{Fapp}). In our ablation we set $p=1$, which, mirroring the per-latent-point form of the Cross-Attention baseline below, yields the encoding
\begin{equation}
\label{FOKernel}
    h(\boldsymbol{y}^j) = \sum_{i=1}^{\bar{N}} k(\boldsymbol{y}^j, \boldsymbol{x}^i)\, v(\boldsymbol{x}^i)
    + \alpha \sum_{l=1}^{M} \sum_{i=1}^{\bar{N}} k(\boldsymbol{y}^j, \boldsymbol{y}^l)\, k(\boldsymbol{y}^l, \boldsymbol{x}^i)\, v(\boldsymbol{x}^i) ,
\end{equation}
where $\{\boldsymbol{y}^l\}_{l=1}^{M}$ are the latent grid points and $\{\boldsymbol{x}^i\}_{i=1}^{\bar{N}}$ the input points. Equivalently, in matrix form this reads $\boldsymbol{V}_G^{(1)} = (\boldsymbol{I} + \alpha \boldsymbol{K}_{GG})\, \boldsymbol{K}_{GP}\boldsymbol{V}_P$, with $\boldsymbol{K}_{GG}$ the kernel Gram matrix on the latent grid; for general $p$ this generalizes to $\boldsymbol{V}_G^{(p)} = (\boldsymbol{I} + \alpha \boldsymbol{K}_{GG} + \cdots + \alpha^{p} \boldsymbol{K}_{GG}^{p})\, \boldsymbol{K}_{GP}\boldsymbol{V}_P$. Thus the $p=1$ setting combines the single-hop kernel aggregation used in prior neural operators with a single extra hop of propagation on the latent grid, and discards all higher-order propagation enabled by the infinite-order resolvent.

\textbf{Cross-Attention.} The second alternative replaces the kernel-based encoder with a standard Cross-Attention mechanism. In this setup, we treat the input coordinates $\boldsymbol{x} \in \Omega$ as queries and the latent grid coordinates $\boldsymbol{z} \in \mathcal{Z}$ as keys; the encoding for a latent point $\boldsymbol{z}_k$ is
\begin{equation}
    h(\boldsymbol{z}_k) = \sum_{i=1}^{N} \text{Softmax} \!\left( \frac{\phi_Q(\boldsymbol{x}_i) \phi_K(\boldsymbol{z}_k)^\top}{\sqrt{d}} \right) \phi_V(u(\boldsymbol{x}_i)),
\end{equation}
where $\phi_Q, \phi_K, \phi_V$ are shallow MLP projections. This replacement eliminates the continuous kernel integration entirely, reducing the mapping to a discrete attention-based interpolation.

\begin{table}[h]
\centering
\caption{Ablation on the encoder mechanism: median relative $L_1$ error ($\%$) of IKNO-Vanilla, IKNO-TP, the first-order kernel variant ($p=1$), and cross-attention. Bold marks the best result per column.}
\label{tab:kernel_vs_attention}
\resizebox{0.8\linewidth}{!}{
\begin{tabular}{lcccc}
\toprule
\textbf{Model} & \textbf{Elasticity} & \textbf{Poisson-Gauss} & \textbf{Wave-C-Sines} & \textbf{NS-PwC} \\ \midrule
IKNO-Vanilla        & 0.93         & 0.34         & 2.86          & 0.83 \\
IKNO-TP             & 0.96 & \textbf{0.26} & \textbf{1.70} & \textbf{0.63}         \\
First-order kernel  & 1.09           & 0.76           & 29.38           & 1.12           \\
Cross-Attention     & \textbf{0.78}          & 1.24          & 85.72         & 62.87         \\
\bottomrule
\end{tabular}
}
\end{table}

The results in Table~\ref{tab:kernel_vs_attention} show that replacing the kernel encoder with Cross-Attention can be competitive on the simpler \textit{Elasticity} benchmark, but it does not provide a robust encoder across tasks. On \textit{Poisson-Gauss} and the two time-dependent benchmarks, IKNO-TP substantially improves over Cross-Attention; the gap is especially large on \textit{Wave-C-Sines} and \textit{NS-PwC}, where Cross-Attention yields much higher errors. The first-order kernel variant further isolates the contribution of the infinite-order expansion: although the $p=1$ truncation preserves a kernel-based latent-grid mapping, it discards higher-order propagation and therefore remains markedly worse than IKNO-TP on the challenging time-dependent problems. Together, these comparisons indicate that the infinite-order kernel mechanism is not merely an architectural detail, but a key component for robustly capturing complex, multi-scale operators.

\section{Computational efficiency}
\label{app:efficiency}

We further investigate the computational cost of IKNO and verify that the Kronecker-structured fast-computation scheme developed in Sec.~\ref{FastComp} preserves the expressivity of the infinite-order formulation without introducing runtime overhead. All measurements in Table~\ref{tab:efficiency} are taken on a single NVIDIA L40S GPU, using the \textit{Wave-C-Sines} benchmark with identical training settings across all methods. We report peak GPU memory and wall-clock time per epoch, and compare against two representative baselines (GAOT and Transolver), as well as an ablation in which the infinite-order resolvent of IKNO is evaluated by explicit $M \times M$ matrix inversion instead of the Kronecker-structured scheme.

Table~\ref{tab:efficiency} shows that IKNO not only matches but outperforms both baselines on efficiency: relative to GAOT, IKNO consumes roughly $24\%$ less GPU memory and runs about $1.7\times$ faster per epoch, while against Transolver it is on par in memory and $\sim\!15\%$ faster. The contrast with the IKNO (direct inverse) ablation is particularly revealing: replacing the Kronecker-structured scheme of Sec.~\ref{FastComp} with direct matrix inversion keeps memory usage roughly unchanged but inflates the per-epoch wall-clock time from $281$s to $1539$s -- a $5.5\times$ slowdown. This empirically confirms the $O(N^{3d})\!\to\!O(d\cdot N^{3})$ complexity reduction predicted in Sec.~\ref{FastComp}, and demonstrates that the practical appeal of the infinite-order kernel formulation relies critically on the Kronecker-based fast computation. The two IKNO variants (Vanilla and TP) exhibit statistically indistinguishable memory and time profiles; we therefore report a single representative IKNO row in the main table and omit a dedicated column per variant.

\section{Evaluation metrics}

To evaluate the performance of our proposed neural operator, we employ different evaluation protocols for non-industrial and industrial datasets to ensure a comprehensive assessment of accuracy and robustness.

\subsection{Non-industrial datasets}
For non-industrial benchmarks, we follow the evaluation protocol of GAOT~\cite{wen2025geometry} and report the median relative $L_1$ error as a percentage. This metric is computed on normalized data to ensure scale-invariance across different physical fields. 

First, let $u_{i,c}$ and $\hat{u}_{i,c}$ denote the ground-truth and predicted fields for test sample $i$ and solution component $c$, respectively. The relative $L_1$ error for that sample-component pair is defined as:
\begin{equation}
    e_{i,c} = \frac{\|u_{i,c} - \hat{u}_{i,c}\|_1}{\|u_{i,c}\|_1}.
\end{equation}

Consistent with GAOT, we take the median of the relative errors to reduce the sensitivity to large outliers. For PDEs with multiple output components, such as velocity and pressure fields, we first compute the median error for each component and then average these component-wise medians, which yields a single scalar score.

\subsection{Temporal evaluation and iterative strategies}
For time-dependent PDEs, we specifically report the performance at the \textbf{final time step}  $T$. Since these systems often require iterative rollouts to reach the terminal state, errors inevitably accumulate over time. Consequently, the final step metric serves as the most demanding and representative indicator of a model's long-term stability and accuracy.

To reach the final state, we implement three distinct inference strategies:
\begin{itemize}
    \item \textbf{Direct:} The model predicts the state at $t=T$ directly from the initial state $u_0$ in a single forward pass (i.e., $\tau = T$).
    \item \textbf{Auto-regressive (AR):} The model evolves the system using the smallest available time step $\Delta t_{min}$. While each individual step may be highly accurate, the large number of iterations required to reach $T$ can lead to significant error accumulation.
    \item \textbf{Star (Strided):} We employ a hybrid approach using a larger stride (the second smallest time step $\Delta t_{next}$). The model iterates using this larger stride until the penultimate state, then determines whether to use a final $\Delta t_{min}$ or $\Delta t_{next}$ step to reach $T$. By reducing the total number of model calls while maintaining temporal resolution, this strategy often achieves the optimal balance between computational efficiency and cumulative accuracy.
\end{itemize}
We evaluate all three inference strategies separately and report the best performance, defined as the minimum error across the methods.

\subsection{Industrial datasets}
For industrial datasets, we adhere to standard regression metrics include Mean Squared Error and Mean Average Error computed on the normalized feature space. 


These metrics provide a direct measure of the absolute error magnitude, which is often prioritized in engineering applications for safety and tolerance analysis.

\section{Visualization of results}
\label{Visualization}

To provide additional qualitative insight into the predictions produced by our IKNO model, we include representative visualizations on the \textit{Poisson-Gauss} and \textit{Wave-C-Sines} benchmarks. The \textit{NASA CRM} visualization is shown in Fig.~\ref{fig:vis_nasa_crm}.

\begin{figure}[h]
    \centering
    \includegraphics[width=0.9\linewidth]{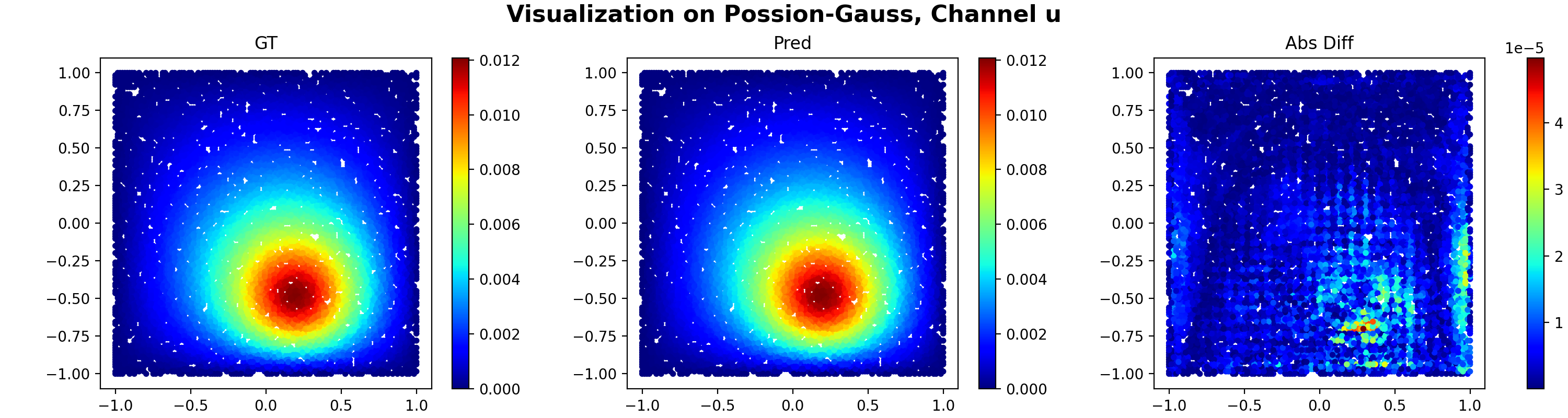}
    \caption{Visualization of IKNO predictions on the \textit{Poisson-Gauss} benchmark.}
    \label{fig:vis_poisson_gauss}
\end{figure}

\begin{figure}[h]
    \centering
    \includegraphics[width=0.9\linewidth]{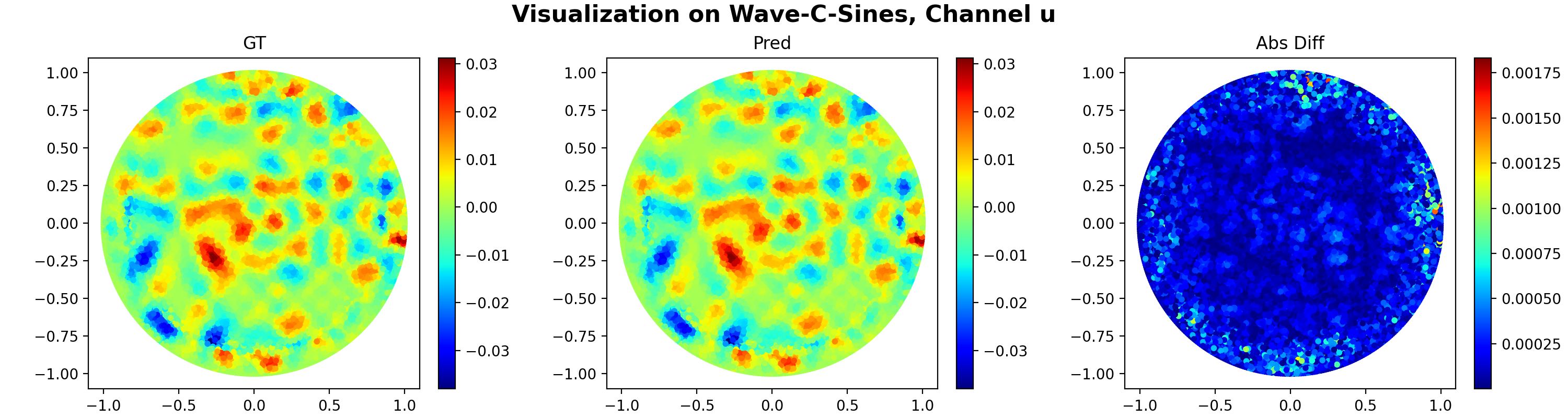}
    \caption{Visualization of IKNO predictions on the \textit{Wave-C-Sines} benchmark.}
    \label{fig:vis_wave_c_sines}
\end{figure}

\end{document}